%% file: main.tex
\documentclass{IOPguidelines/iopjournal}

\usepackage{amsmath} 
\usepackage{xcolor}
\usepackage[T1]{fontenc}   
\usepackage{lmodern}       
\usepackage{amsfonts}
\usepackage{graphicx}
\usepackage{booktabs}
\usepackage{pgfplots}
\pgfplotsset{compat=1.18}
\usepackage{booktabs,tabularx}
\usepackage{hyperref} 
\usepackage{siunitx}
\usepackage{cleveref}
\usepackage{threeparttable}
\usepgfplotslibrary{fillbetween}
\usetikzlibrary{intersections}
\usepackage{newtxtext,newtxmath}
\usepackage{graphicx}
\usepackage{booktabs}
\usepackage{multirow}
\usepackage{tabularx}
\usepackage{array}
\usepackage{xparse}
\usepackage{caption}
\usepackage{titlesec} 
\usepackage{subcaption}
\usepackage{float}
\usepackage{titlesec}
\usetikzlibrary{pgfplots.groupplots}
\usepackage[backend=biber,style=vancouver,sorting=none]{biblatex}
\graphicspath{{./}{IOP guidelines/}}

\titleformat{\paragraph}[runin]
  {\normalfont\itshape} 
  {\theparagraph}{1em}{}[:]
\titlespacing*{\paragraph}{0pt}{3.25ex plus 1ex minus .2ex}{1em}

\definecolor{pomcBlue}{RGB}{120,180,255}
\definecolor{hxExplore}{RGB}{115,179,158}
\definecolor{hxEmMin}{RGB}{245,170,110}
\definecolor{hxDynEm}{RGB}{165,195,230}
\definecolor{hxProfMax}{RGB}{215,160,210}
\definecolor{hxDynProf}{RGB}{185,200,120}

\definecolor{lightblue}{rgb}{0.55,0.75,0.95}
\definecolor{lightred}{rgb}{1.00,0.60,0.60}
\definecolor{lightgreen}{rgb}{0.60,0.85,0.60}

\begin{document}

\title{Optimizing Lithium Production Decisions under Geological, Demand, and Pricing Uncertainties: A POMDP Framework for Multi-Objective Decision Making}

\author{
  Anna C. Edmonds$^{*1}$\orcid{0009-0008-6528-9067},
  Mansur M.\ Arief$^{*1,3}$\orcid{0000-0002-4636-3451},
  Robert J.\ Moss$^{*1}$\orcid{0000-0003-2403-454X},
  Mykel J.\ Kochenderfer$^{1,2}$\orcid{0000-0002-7238-9663},
  and Jef Caers$^{3}$\orcid{0000-0003-0262-9905}
}

\affil{$^1$Computer Science Department, Stanford University, Stanford, CA, USA}

\affil{$^2$Aeronautics and Astronautics Department, Stanford University, Stanford, CA, USA}

\affil{$^3$Earth and Planetary Sciences Department, Stanford University, Stanford, CA, USA}

\affil{$^*$Author to whom any correspondence should be addressed.}

\email{\{edmondsa, ariefm, mossr\}@stanford.edu}

\keywords{Lithium mining, critical minerals, mine planning under uncertainty, partially observable Markov decision process (POMDP), multi-objective optimization, price and demand uncertainty, direct lithium extraction (DLE), hard-rock lithium mining}

\begin{abstract}
Decision-making in lithium production is challenging, whether from an investor’s perspective or a strategic production standpoint. Determining which mines to open and when to open them involves not only geological and price uncertainties, but also complexities around the choice of extraction method, from direct lithium extraction to hard rock mining. Prior work explored models of this problem and different methods to optimize mining decisions; these models did not account for uncertainty in pricing, uncertainty in demand, or different mining technologies to extract lithium. Incorporating different pricing models and extraction technology into these models enables more robust strategies for determining not only when and where to open a mine, but also which method of production to pursue. We frame the problem as a partially observable Markov decision process (POMDP) and solve using belief state planning methods to get optimal decision making. In our study, we show that POMDP solvers outperform human-inspired heuristics by dynamically adapting to shifting lithium price regimes (static, linear, exponential, and stochastic) through belief-state planning and explicit uncertainty management. By optimally sequencing exploration, production, and technology choice, the framework achieves higher demand fulfillment and more balanced economic–environmental outcomes over the projects lifetime in all different pricing and deposit scenarios.

\end{abstract}
\section{Introduction}


Lithium (Li) is one of the most essential raw materials for the clean energy transition as it is essential in large-scale batteries for the grid and electric vehicles \cite{ATTILIO2025124688, iea_netzero_2021}. The demand for lithium is growing the fastest of all other major minerals \cite{iea_criticalminerals_2021, Xu2020, giurco2019} with 85\% of this demand coming from battery sector \cite{Mehdi2024}. Lithium had a demand of 165 kt in 2023 and according to the IEA it is predicted that the demand will rise to 1326 kt by 2040, that is an increase of 706\%  \cite{iea_lithium_2024, iea_lithium_demand_chart_2024}.

Meeting this rapidly growing lithium demand poses significant challenges on the supply side, as production remains geographically concentrated and constrained by the limited availability of economically viable deposits. Hard-rock (spodumene) mining, primarily in Australia, currently supplies around 50--60\% of global lithium, while brine extraction operations in South America’s ``Lithium Triangle'' (Chile, Argentina, and Bolivia) contribute most of the remainder \cite{kesler2012global}.  The narrow geographic distribution of high-grade lithium resources introduces supply vulnerabilities, since production capacity cannot be expanded quickly enough to meet sudden demand surges. Mine development typically requires 5--10 years from discovery to operation \cite{bottomley2003lithium}, creating long lead times that limit supply responsiveness. These challenges are compounded by environmental and social constraints, including water scarcity and carbon emissions in brine-rich regions, as well as increasingly stringent permitting and regulatory requirements. Together, these factors constrain the ability of both hard-rock and brine extraction methods to scale in pace with global demand growth at a correct time. \cite{rentier2024lithium}.


The supply-demand gap has arguably caused lithium carbonate spot prices to exhibit extreme volatility in recent years, reflecting the cyclical dynamics of the global EV and energy storage markets. Due to tightening supply conditions and surging electric-vehicle demand during 2021–2022, lithium prices rose by more than 150\% \cite{nrcan_lithiumfacts_2025}, reaching unprecedented highs. They subsequently fell by over 80\% in 2023 amid growing concerns about oversupply, the phase-out of Chinese EV subsidies, and weaker-than-anticipated electric-vehicle demand \cite{Mehdi2024}.  The IEA notes that lithium is experiencing its highest price volatility among focus minerals \cite{iea_lithium_2024}, which many believe is due to the complex market dynamics between the supply and demand of lithium. It has been shown that anticipating future price trends of lithium is important to avoid supply shortages and unstable supply chains \cite{li_uncertainty_management, resourpol_bubbles_2023} as it may lead to market inefficiencies and bad resource allocation \cite{resourpol_bubbles_2023}.

Furthermore, recent advances in extraction technologies are reshaping the lithium supply landscape, adding even greater uncertainty to the optimal pathway for production. While traditional production has relied on hard-rock and brine-based operations, new methods such as Direct Lithium Extraction (DLE) are emerging as potentially transformative for brine resources \cite{FARAHBAKHSH2024117249}. DLE technologies aim to increase lithium recovery rates and reduce water and land intensity by selectively extracting lithium ions from brines, most commonly by using chemical sorbents or membranes \cite{goldman_dle_2023}. DLE offers faster processing times and a smaller environmental footprint; however, its commercial scalability and cost competitiveness remain uncertain as it is a relatively newer technology. In contrast, hard-rock operations require higher upfront capital expenditure (CAPEX) and greater energy requirements but generally benefit from lower operating expenditures (OPEX) (refer to Appendix \ref{appendix:dle_vs_hardrock} for CAPEX and OPEX comparison). Despite these trade-offs, most models optimizing lithium mining ~\cite{arxiv_geouncertainty_2025} have not incorporated technological differentiation, treating hard rock mining and DLE the same. Our framework explicitly distinguishes between extraction technologies because deposit size, recovery rate, costs, and emissions vary across technologies, so the preferred option can shift with changes in demand, prices, and other uncertainties over time.


The main challenge that we aim to address is how to optimize lithium production decisions for investors and mining companies, under these geologic, economic, and technological uncertainty. These intertwined uncertainties place mounting pressure on suppliers to achieve both environmental and economic viability~\cite{arxiv_geouncertainty_2025}; thus, delivering lithium with a low life-cycle carbon footprint comparable to or better than fossil-fuel alternatives, while maintaining profitability in an increasingly competitive and dynamic energy market~\cite{khakmardan2023comparative}. Therefore, strategic questions about what to develop, when to commit capital, and how to meet demand must be addressed under incomplete information and dynamic markets~\cite{HALKES2024107554} .

Previous research examining mining investment under uncertainty has often relied on econometric and decision-analytic frameworks to analyze the timing of mine development decisions. Real options theory has been widely used to interpret how firms respond to commodity price uncertainty and the flexibility to open or close mines \cite{rfs_realoptions, HAQUE2014115}. Empirical studies have also applied econometric methods such as logistic or ordinal regression to analyze large datasets of mining projects and identify factors associated with the progression of exploration projects to production \cite{resourpol_lithium_projects_2024}. While these approaches provide valuable insights into industry-level investment behavior, they are primarily descriptive and do not explicitly model sequential decision-making under partial information.

To address this challenge, this work employs a partially observable Markov decision process (POMDP) framework. POMDPs are well-suited to the lithium supply context because they explicitly model situations in which the true system state (such as deposit quality, recoverable reserves, or future market conditions) is only partially known. Rather than assuming perfect information, decision-makers act on probabilistic \textit{beliefs} about the underlying state and update these beliefs as new geological or market data become available.

Our framework models price and demand uncertainty through multiple dynamics (static, linear, exponential, and stochastic) and evaluates each decision strategy against historical price data to ensure robustness across market regimes. It also captures key operational decisions: when to explore or expand, how to allocate capital among geographically and geologically diverse deposits, and which extraction technology (DLE or conventional hard-rock mining) offers the best trade-off between cost, efficiency, and environmental impact. These choices carry high capital intensity and long lead times, making premature or misinformed investments extremely costly. Traditional optimization or deterministic forecasting approaches fail to capture the coupled, evolving nature of these systems, where decisions made today influence both the information available and the feasible options tomorrow \cite{arief2025managing}. In contrast, POMDPs balance exploration (investing in information through drilling, pilots, or market observation) against exploitation (committing to production once sufficient confidence is gained). As shown in the results, this framework provides a solid foundation for decisions that balance economic performance with environmental considerations in a rapidly changing resource and political landscape.

\section{Methodology}
This section outlines the computational framework used to model lithium extraction as a sequential decision-making problem under uncertainty. We describe the components of a POMDP are, how our problem is formulated as a POMDP, and how it is solved. Our work is open-sourced and can be accessed here: \url{https://github.com/sisl/LiPOMDP.jl}

\subsection{Partially Observable Markov Decision Processes}
To address the multiple uncertainties of lithium geology, price dynamics, and market demand described above, this work models mine planning of multiple mining operations as a partially observable Markov decision process (POMDP). Unlike a fully observable Markov decision process (MDP), a POMDP captures decision-making when state variables, such as in-situ deposit volumes, recoverable lithium, or future market prices, cannot be directly observed.

Formally, the POMDP is represented as a seven-tuple 
$\langle \mathcal{S}, \mathcal{A}, \mathcal{O}, T, R, \Omega, \gamma \rangle$, where \( \mathcal{S} \) is the set of possible states of the model, \( \mathcal{A} \) is the set of available actions, \( \mathcal{O} \) represents the observation space, \( T(s' \mid s, a) \) defines the state transition dynamics, \( R(s, a) \) specifies the reward function based on the state and action, \( \Omega(o \mid a, s') \) defines the observation model, and \( \gamma \) is the discount factor which controls how much near term vs. the far future rewards that are valued \cite{kochenderfer2022algorithms}.

At each time step, the agent selects an action \( a_t \in \mathcal{A} \) based on its current belief about the state \( b_t \), which is a distribution over the unobserved state. The environment then transitions to a new unobserved state based on the action it takes \( s_{t+1} \sim T(s' \mid s_t, a_t) \), which produces a reward \( R(s_t, a_t) \), and generates an observation \( o_{t+1} \sim \Omega(o \mid a_t, s_{t+1}) \). Then the agent updates its belief using the observation to form \( b_{t+1} \), closing the loop between belief about the state and decision-making on what action to take. Through this recursive interaction among \( \mathcal{S}, \mathcal{A}, T, \Omega, \) and \( R \), the POMDP captures the coupled dynamics of uncertainty and planning, where it can determine a policy that maximizes the expected discounted sum of future rewards, which maps a belief to an action \cite{kochenderfer2022algorithms, kozlova2024sensitivity, ross2008online}.

\subsection{Application of Lithium Mining}
The POMDP framework provides an ideal structure for modeling mine-planning decisions under uncertainty, as it captures the interplay between incomplete information, dynamic planning, and multi-objective trade-offs that characterize lithium extraction projects. This model framework enables sequential decision-making under multiple layers of uncertainty, geological, economic, and environmental, typical in long-horizon mining investments. By continuously updating its beliefs and optimizing over future expected rewards, the agent learns to balance exploration and exploitation intelligently: exploring to reduce geological uncertainty and uncover profitable, low-emission opportunities, while exploiting at the right time based on price and demand across the project’s lifetime.

The model incorporates both DLE and hard-rock spodumene mining, requiring the agent to decide which technology to deploy first based on factors such as capital and operating costs, carbon intensity, recovery efficiency, and technological uncertainty (see Appendix \ref{appendix:params}). The decision-making entity representing the agent (for example, an investment firm or mining company) must determine whether, when, and where to initiate extraction, and which technology to employ under uncertainty. In the following sections, we describe the key components of our lithium POMDP, which together simulate the decision-making environment of firms seeking to optimize both profit and emissions outcomes under uncertainty in the lithium market.The parameter values used in this study are illustrative and can be substituted with project-specific inputs.

\subsection{States}
Each state \( s_t \) encapsulates both the physical and economic conditions of the lithium mines. Physically, the state represents the amount of lithium available at each potential mining site, measured in tonnes of lithium metal. To keep the model focused and manageable, we consider four deposits in total: two using DLE and two using hard-rock mining. Direct lithium extraction sites are smaller, cleaner, and lower in capital costs, while hard-rock sites are larger and more carbon-intensive \cite{khakmardan2023comparative}. Refer to Appendix \ref{appendix:params} for a full overview of the parameters used in the model.

The state also includes the lifecycle of the mine indicators that capture whether each mine is (1) unopened, (2) opened and producing, (3) depleted, or (4) restored following closure \cite{resourpol_lithium_projects_2024}. Economically, the state tracks the lithium carbonate equivalent (LCE) market price and the entity’s projected demand, defined as 30\% of the market share of global production projections that the agent strives to reach sequentially. Refer to Appendix~\ref{appendix:historic_price} for the historic price and global lithium demand. Together, these variables define the state of the environment in which mining decisions are made, linking resource availability with market opportunity and environmental and economic trade-offs.

\subsection{Observations}
Because in-situ deposits and recoverable volumes cannot be directly observed, the agent relies on noisy observations that depend on its actions to capture geological, price, and market uncertainty. These observations generated over time provide the agent with clues of the true underlying states (hence partially observable), updating its belief of the current state. Each observation \( o_t \) consists of three components:

\begin{enumerate}
    \item \textbf{Deposit estimates for each site}, sampled from Normal distributions whose variance depends on the action: low noise when exploring (high fidelity), moderate when operating (learning from production), and high when idle (limited information) (see Appendix~\ref{appendix:observation_noise}). 
    
    \item \textbf{Observed market price}, sampled from a Normal distribution centered on the true price with variance based on historical volatility.
    
    \item \textbf{Observed demand}, sampled from a Normal distribution centered on the true company demand with variance (see Appendix~\ref{appendix:historic_price}).
\end{enumerate}

This structure reflects how information quality in lithium production depends on actions: exploration and operation reduce uncertainty, while inaction maintains ignorance about both subsurface and market conditions.

\subsection{Actions}
The action space represents the discrete set of lifecycle decisions available to the mining firm, which was simplified from what is done in industry \cite{resourpol_lithium_projects_2024, etde_20847437}. Each of the four sites can take one of four actions, giving an initial action space of \( 4^4 = 256 \) combinations. This space quickly shrinks as sites advance through their stages; for example, once a site is opened, the range of available actions for that site becomes restricted, as it can no longer explore, reducing the total number of possible joint actions over time.

\begin{enumerate}
    \item \textbf{Explore:} Incurs an exploration cost but improves knowledge of the deposit distribution.
    \item \textbf{Open:} Triggers CAPEX, activates production, and transitions the site to the producing state.
    \item \textbf{Restore:} Represents decommissioning and environmental rehabilitation of a depleted site, incurring a one-time cost and emissions.
    \item \textbf{Do nothing:} Leaves the site unchanged for that period. If the mine has been opened and not restored, it will incur an OPEX and emission cost per tonnes of Li produced. 
\end{enumerate}

\subsection{Transition}

The transition \( T(s_{t+1} \mid s_t, a_t) \) describes how the system evolves to a new state after actions are taken based on the current state \cite{kochenderfer2022algorithms}. Time is discretized in one-year increments, consistent with standard mine-planning horizons, so each decision epoch represents a single operational year. Over each yearly step, the model updates physical, economic, and environmental variables, capturing production, depletion, and price fluctuations as the system transitions to a new state. The deposits deplete deterministically as lithium is extracted, while price and demand evolve stochastically.

\subsubsection{Production}
Each open mine produces up to its technical capacity, which for our study is set to \( 3{,}500 \) tons of Li, capped by the remaining deposit and the company’s current-year demand, defined as \( 30\% \) of the historical global production that year (see Appendix \ref{appendix:params} for parameters). Sellable output is computed as \( \mathrm{LCE}_i = \min(\text{capacity}, D_{t,i}) \times \text{recovery}_i \times \text{LCE per Li} \). The in-situ deposit is then reduced by the amount mined, and when \( D_{t,i} \approx 0 \), the mine is marked as depleted.

\subsubsection{Pricing}
For each pricing model, we wrote an equation for how the price changes over time and then fitted the constants in that equation to the historical lithium price data from 1994 to 2024. This allowed each model to reflect the general shape and level of real prices rather than being created from arbitrary values. In \cref{appendix:price_equation_constants} lists the fitted constants and the starting price $p_0$ used for each model. All prices are LCE, which is 5.35 times the price of lithium metal. 

\paragraph{Static Pricing}  
Under the static pricing regime, the lithium price fluctuates around a constant baseline over time. We fix the baseline price at $p_0$ and model prices as $p_t = p_0 + \epsilon_t$, where the noise term follows $\epsilon_t \sim \mathcal{N}\bigl(0,(0.05\,p_0)^2\bigr)$. This corresponds to a standard deviation equal to 5\% of the baseline price, capturing minor market variability while preserving the underlying assumption of static behavior.

\paragraph{Linear Pricing}
Linear pricing introduces a deterministic time-based trend in lithium prices, expressed as $p_t^{\mathrm{det}} = p_0 + \alpha t$, where $p_0$ is the starting price and $\alpha$ is the linear slope. To incorporate real-world volatility, we model the realized price as $p_t = p_t^{\mathrm{det}} + \epsilon_t$ with proportional noise $\epsilon_t \sim \mathcal{N}\bigl(0,(0.1\,p_t^{\mathrm{det}})^2\bigr)$, ensuring that the variance scales with the current price level. This formulation simulates markets where prices follow a predictable trend but still can exhibit dependent fluctuations.

\paragraph{Exponential Pricing}
Exponential pricing captures accelerating price trajectories resulting from compounding demand or inflationary pressures. The deterministic evolution is given by $p_t^{\mathrm{det}} = p_0 e^{\lambda t}$, where $\lambda$ is the exponential growth rate. As in the linear model, we introduce price-level-proportional volatility by setting $p_t = p_t^{\mathrm{det}} + \epsilon_t$ with $\epsilon_t \sim \mathcal{N}\bigl(0,(0.1\,p_t^{\mathrm{det}})^2\bigr)$. This structure reflects markets in which both the expected price and the volatility increase multiplicatively over time.

\paragraph{Geometric Brownian Motion}
Lithium prices were also modeled using a Geometric Brownian Motion (GBM)~\cite{stojkoski2020generalised}, which assumes continuous stochastic growth governed by a drift term $\mu$ and volatility $\sigma$. In discrete time with step size $\Delta t$, this corresponds to the update rule $S_{t+\Delta t} = S_t \exp\!\bigl((\mu - \tfrac{1}{2}\sigma^2)\Delta t + \sigma\,\Delta W\bigr)$, where $\Delta W \sim \mathcal{N}(0,\Delta t)$ represents a Wiener increment. To estimate $\mu$ and $\sigma$, we computed annual log-returns $r_t = \ln(S_t / S_{t-1})$ using 30 years of observed lithium spot prices and applied a 5\% winsorization to mitigate the influence of extreme events such as the 2022 price spike. The resulting estimates, $\mu \approx 0.033$ and $\sigma \approx 0.238$, indicate modest long-term growth with moderate volatility and provide stable inputs for stochastic simulation and real-options analysis under uncertainty.

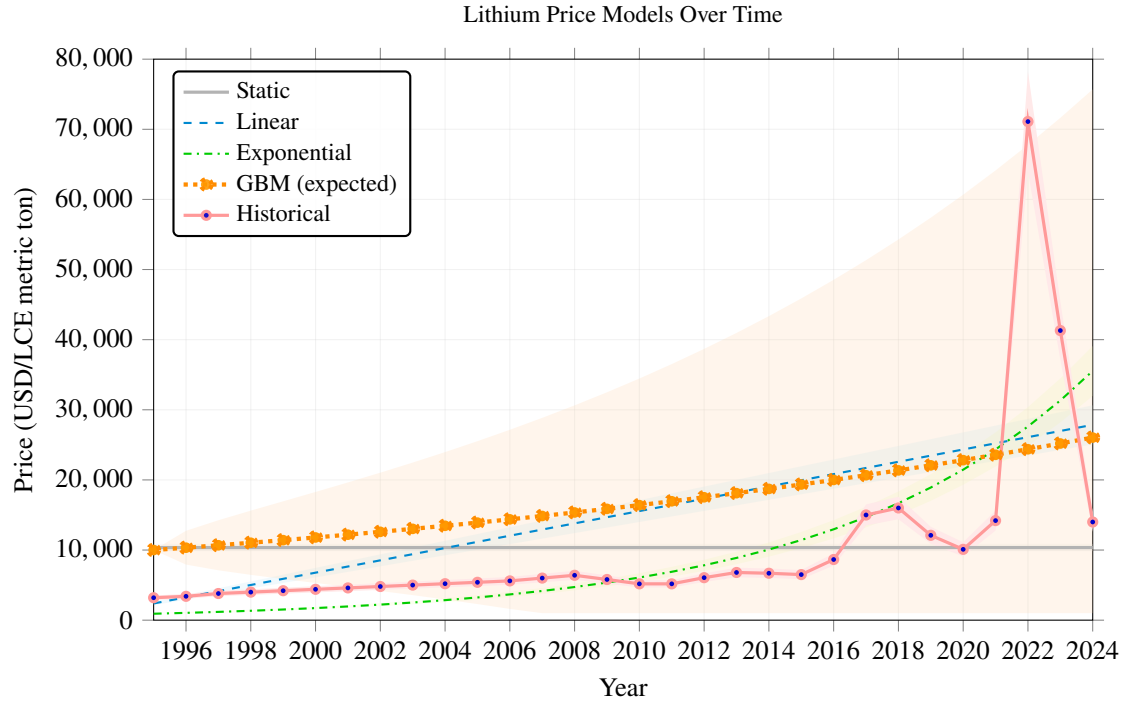
\begin{figure}[H]
    \centering
    \input{figures/pricing_models_graph} 
    \caption{Pricing Models with their Standard Deviation over Time period}
    \label{fig:pricing_models}
\end{figure}

Illustrated on \Cref{fig:pricing_models} are the five pricing models used in this study, showing each fitted trajectory alongside the historical price data \cite{usgs_lithium_stats_2025, statista}, which rose sharply in 2021 and 2022 before declining in the following years.

\subsection{Reward}
The reward function represents the balance between economic profit and environmental cost, guiding decisions about which type of mine to open, when to operate, and how to plan under uncertainty. Each action carries financial and environmental consequences, and the reward combines both into a single objective.

When a mine is open, it automatically produces lithium up to its technical capacity, which is the same for both DLE and hard-rock mines, but is limited by the remaining deposit and the company’s demand in that period. The extracted lithium is converted to LCE using the recovery efficiency of each technology, dependent on whether the site uses DLE or hard-rock mining (see Appendix \ref{appendix:params}), and a fixed Li-to-LCE conversion factor of 5.323~\cite{MIATTO2020105034}. Company demand resets each year, ensuring that production decisions are evaluated annually. (see Appendix~\ref{appendix:historic_price} for parameters).

Revenue (positive rewards) for each period is calculated as the market price of LCE (which depends on the chosen price model, i.e. static, linear, exponential, Geometric Brownian motion, or historical) multiplied by the amount sold. The quantity sold from each open site is computed as the recovered lithium output converted to LCE and is capped simultaneously by the site’s technical capacity, the remaining in-situ deposit of the mine, and the firm’s remaining demand in that period.

Costs (negative rewards) include several components. Exploration costs are incurred whenever a site is explored. When a new site is opened, the company pays a one-time CAPEX, followed by OPEX proportional to the amount of LCE produced. Once the mine is depleted, a restoration cost is incurred. CAPEX, OPEX, and restoration costs differ between DLE and hard-rock technologies, making it essential for the POMDP to plan which type of mine to open and when, given the price and demand each year. These terms together represent the full economic cost of operating in a given year (see \cref{appendix:params,appendix:dle_vs_hardrock} for parameters).

Emissions are modeled in a similar structure as negative rewards. Every operating mine has an emissions per tonne of LCE produced depending on if its DLE or hard-rock mining. ~\cite{khakmardan2023comparative}. All emissions per tonne are then multiplied by an equivalent cost of \$190 per tonne of CO$_2$ using the social cost of carbon published by the EPA under the Biden administration \cite{epa2023scghg}, ensuring that environmental impact enters the same decision framework as financial profit.

The final reward combines both components:
\[
R(s_t, a_t) = \frac{\alpha \, (\text{revenue}_t - \text{costs}_t) + (1 - \alpha)\, (-\text{CO}_2{}_t \times C_{\text{CO}_2})}{P_{\text{reward\_scale}}}
\]
where $\alpha \in [0, 1]$ controls the trade-off between profit and $C_{CO_2}$ is the emission cost multiplied by the total carbon dioxide $CO_{2t}$ produced at time \textit{t}.

\subsection{Belief State Updating}
Because the agent cannot directly observe the true system state, it maintains a belief state $b_t$---a probabilistic estimate of the underlying conditions of deposit size, market demand, and lithium price, each of which carries uncertainty. This belief is modeled as a distribution over possible states and is iteratively updated as the agent takes actions and receives new observations.

For each site, these uncertain variables are modeled as Gaussian (Normal) posterior distributions conditioned on past observations. When an action $a_t$ is taken and a new observation $o_{t+1}$ is received, the belief is updated through a Kalman filter. At its core, this filter performs a recursive application of Bayes’ rule, updating the mean and variance of the prior belief state $p(s_t \mid o_{1:t-1})$ given the new observation noise. The resulting posterior belief state $p(s_t \mid o_{1:t})$ reflects an improved estimate of the true state, with reduced variance as the agent accumulates more information, mirroring the process of learning over time~\cite{ristic_beyond_2004, chong2009partially}. The noise level depends on the action: it is low during exploration (high-quality geological data), moderate during production (operational data), and high when idle (limited feedback). 

\subsection{Solving the POMDP}

Solving a POMDP involves finding an optimal policy $\pi^*(b)$ that maps each belief state $b_t$ to an action $a_t$ maximizing the expected discounted cumulative reward $\mathbb{E}\!\left[\sum_{t=0}^{\infty} \gamma^t R(s_t, a_t)\right]$, where $\gamma$ captures the agent’s time preference between short- and long-term outcomes. In our model, this corresponds to selecting actions that optimally trade off profit and emissions, with relative weighting determined by $\alpha$.

Two main approaches exist for solving POMDPs: \textit{offline} and \textit{online} methods~\cite{ross2008online, juliaPOMDPdocs}. Offline solvers attempt to pre-compute a policy for all possible beliefs before execution. This approach works well for small or discrete problems but becomes computationally infeasible in large or continuous domains. Online solvers, in contrast, plan in real time from the current belief. Rather than computing an entire policy in advance, they simulate potential future trajectories from the agent’s current belief to estimate which immediate action is most promising. Online methods construct belief trees rooted in the current belief \cite{WANG2022104266}, expanding branches based on possible actions, resulting observations, and updated beliefs up to a finite depth. This allows the agent to plan adaptively as new information arrives, focusing computation only where it matters most.

We employ the POMCPOW (Partially Observable Monte Carlo Planning with Observation Widening) algorithm~\cite{sunberg2018online}, an online planner that efficiently handles continuous, uncertain environments. POMCPOW uses Monte Carlo simulations to explore many possible outcomes from the current belief, progressively widening both the action and observation spaces. This approach makes it particularly suitable for the lithium mining problem, where the agent must continuously update its strategy under uncertainty rather than rely on a static, precomputed policy \cite{ross2008online}.

\subsection{Experiments}
We assess whether the online POMDP planner (POMCPOW) can surpass rule-based heuristics designed to reflect the decision processes of human actors, such as investors or mining companies, when operating under uncertainty.  To benchmark performance, POMCPOW is compared against seven heuristic baselines designed to represent different kinds of human-like decision frameworks. These heuristics used are as follows:
\begin{enumerate}
    \item \textbf{Random}: For each unopened site, randomly choose among \{\textit{do nothing}, \textit{explore}, \textit{open}\}; always restore when depleted.
    \item \textbf{Explore-Only}: Continuously explores each mine to decrease geological uncertainty, without ever opening for production.
    \item \textbf{Single Action Profit Maximizer (one timestep)}: At the first time step, opens the site with the highest expected lifetime profit (using the average lifetime price from the active price model, realistic demand, and depletion limits). 
    \item \textbf{Single Action Emission Minimizer (one timestep)}: Opens the site with the lowest total emissions, summing all expected emission costs over the remaining horizon.
    \item \textbf{Dynamic Profit Maximizer (per-step)}: At each step, evaluates unopened sites for incremental lifetime profit given current beliefs, demand limits, price trajectory, and emission costs. Opens the best site, selectively explores promising alternatives, and restores when depleted.
    \item \textbf{Dynamic Emission Minimizer (per-step)}: Opens new sites only if current capacity cannot meet demand. Chooses the site(s) adding the least total emissions to close the gap; otherwise, explores selectively and restores when depleted.
    \item \textbf{One-Step Lookahead (OSL)}: At each step, enumerates feasible joint actions, simulates rollouts, and selects the action with the highest near-term expected return. This represents a myopic look-ahead without belief tree widening.
\end{enumerate}

POMCPOW is tuned using 25 training simulations to calibrate planning hyperparameters (e.g., tree depth, query budget, and widening constants) and evaluated on 50 held-out test simulations using distinct random seeds to avoid overfitting. Heuristic baselines are non-learning policies; thus, they are tested only on the 50 evaluation runs. We also used POMCPOW with a depth of 1 as a baseline to assess how POMCPOW performs with minimal lookahead.

Because market volatility creates significant uncertainty in investment timing and the balance between demand and supply~\cite{resourpol_lithium_projects_2024}, we plan policies under static, linear, exponential, and stochastic (GBM-based) price models and then evaluate the actions against historical prices (refer to Appendix~\ref{appendix:historic_price}). This approach tests how robust the POMDP framework is when price dynamics are uncertain and shows how well it can plan under any pricing environment.  This setup mimics an investor who optimizes decisions under a particular price model and later discovers how those decisions would have performed under the actual historical price trajectory.

Each policy is then evaluated across multiple performance metrics, including discounted reward, total profit, total emissions and met demand, for varying values of the trade-off parameter $\alpha$. This parameter controls the relative importance of economic versus environmental objectives within the reward function, allowing it to be tuned to reflect the strategic priorities of the decision-maker. For example, an investment firm focused on financial returns might assign a higher $\alpha$, emphasizing profit maximization, whereas a company prioritizing sustainability and regulatory compliance might choose a lower $\alpha$, emphasizing emission reduction. By varying $\alpha$, the model generates a spectrum of strategies along the profit–emissions frontier (see \cref{fig:pareto_curve}), illustrating how decision preferences shape optimal behavior under uncertainty.

In addition to these primary objectives, we also evaluate policies on secondary operational metrics including the percentage of demand met, total LCE produced, time to first production, number of mines opened, and total exploration expenditure. Although the policy is trained to optimize only the reward components of profit and emissions, these additional measures provide a broader view of performance, revealing how different strategies behave in practice and how effectively the POMCPOW planner balances profitability, sustainability, and operational resilience over the project lifetime in comparison to the heuristics.

\section{Results and Discussion}

We consider this setting from the perspective of a decision-maker (e.g., a mining company or investor) making sequential investment decisions under uncertainty from 1994 to adapting through until 2024. At each time step, the decision-maker must determine whether to explore, develop, delay, or restore lithium assets, and which extraction technology (DLE or hard rock) to deploy, without full knowledge of subsurface resources or future price and demand trajectories. Across all pricing models (shown in \cref{tab:p1e6_alpha05,tab:p2e6_alpha_0.5,tab:alpha05_results,tab:p4e6_alpha05,tab:p6e6_alpha05}), POMCPOW consistently achieves the highest or near-highest discounted reward, often with competitive emissions and strong demand satisfaction, demonstrating its superior ability to balance exploration, profitability, and environmental considerations. The discounted reward metric captures both immediate and long-term outcomes, and POMCPOW’s advantage indicates its capacity to make forward-looking decisions that account for uncertainty in geology, price, and demand. At each step it simulates many possible futures under the assumed price model, updates its beliefs with new exploration and production data, and explicitly optimizes the alpha-weighted profit–emissions objective. The heuristics, by contrast, apply fixed rules (e.g. lifetime profit or emission intensity thresholds) to the current state without lookahead or explicit uncertainty handling. As a result, POMCPOW can better time mine openings and restorations in the face of uncertain geology, demand, and price trajectories, particularly in non-static price environments. This pattern holds despite the different price dynamics and noise structures implied by each model, indicating that the planner’s belief-aware look-ahead is robust to both smooth, static and linear, and more volatile, exponential and stochastic (GBM), markets. 


Belief uncertainty is central to how POMCPOW learns and adapts under incomplete information. As shown in the belief-uncertainty plots (Figures~\ref{fig:deposit_uncertainty_grid}), the model continuously updates its beliefs on the geological deposits as new data are revealed through exploration and production. The belief uncertainty plots illustrate the dynamics of information acquisition under different policies. The Explore-Only policy gradually reduces uncertainty through always exploring. Conversely, the Profit Maximizer acts too aggressively, committing capital before uncertainty is sufficiently resolved. POMCPOW reduces uncertainty not by exploring randomly or committing to opening a mine too early, but through sequential, information-driven decision-making. Each action improves the model’s understanding of the underlying deposits, which in turn informs the next step. By continuously updating its belief of the system’s state and adjusting its strategy accordingly, POMCPOW learns when to explore, when to open a mine, and when to hold back and do nothing. This structured reasoning enables it to make more informed choices over time, leading to higher discounted rewards, steadier emissions performance, and stronger long-term profitability.

\input{figures/belief_uncertainty_grid}

\begin{figure}[H]
    \centering   
    \includegraphics[width=\linewidth]{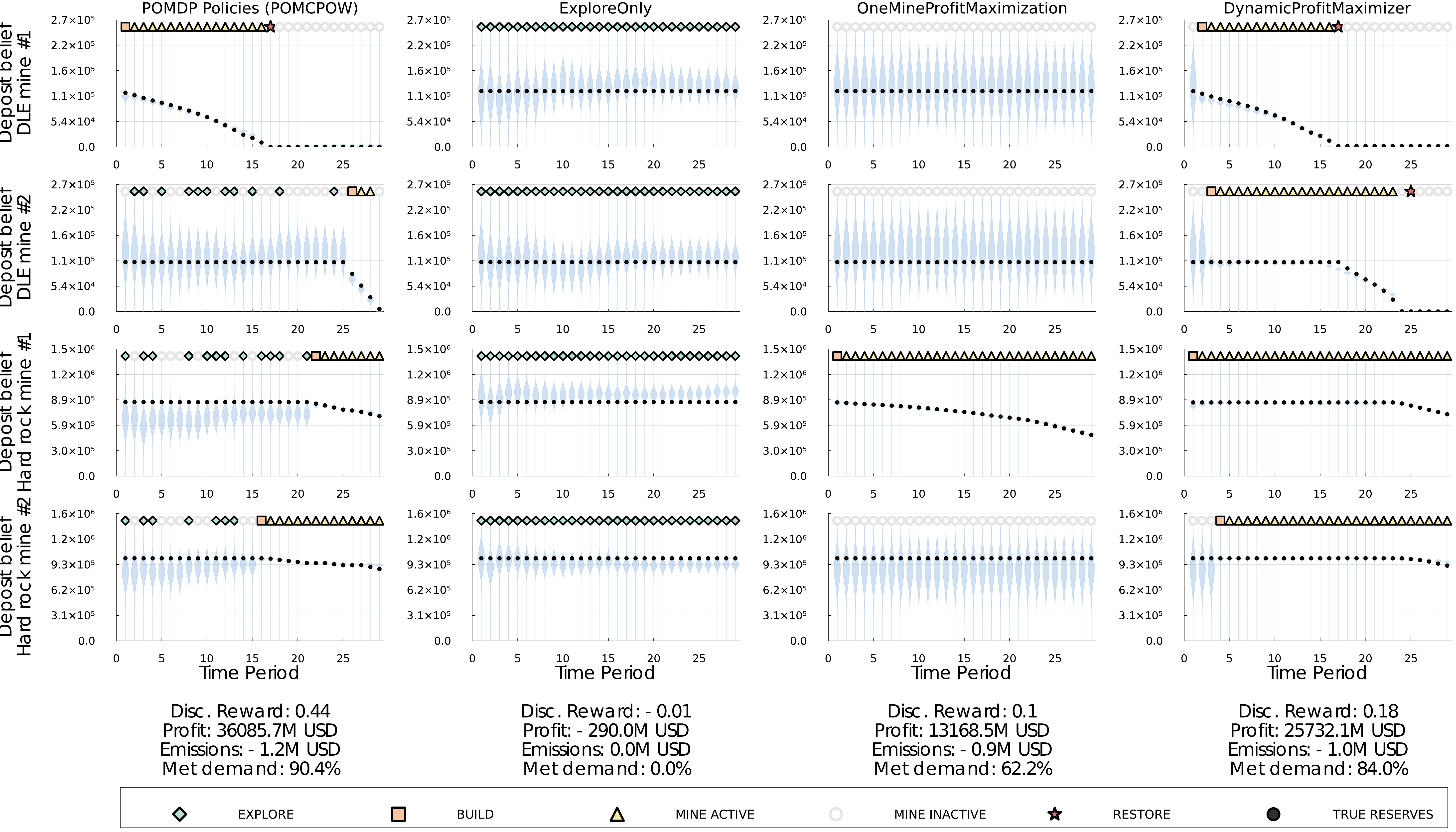}
    \caption{Action timeline over the 29-year horizon showing belief uncertainty and action selection for four policies. All policies plan under the Geometric Brownian Motion pricing model and are evaluated under historical pricing across four lithium deposits in a representative seed run.}
    \label{fig:action_timeline}
\end{figure}

This sequential decision-making structure, guided by systematic reduction in belief uncertainty, is further illustrated in \cref{fig:action_timeline}, where we compare POMCPOW’s action sequence to the heuristic policies. At the beginning of the horizon, POMCPOW immediately opens DLE mine \#1. This choice reflects its lower capital cost, lower emissions intensity, and higher recovery rate, which together provide early positive cash flow with comparatively limited upfront risk. Under discounting, early revenue is more valuable, making the lower-capex option optimal while uncertainty over total reserves remains high. Simultaneously, the policy allocates exploration actions to the remaining deposits. These actions reduce belief uncertainty regarding reserve sizes without requiring irreversible capital commitments. As the belief distributions narrow and align closer to true reserves and both demand and price increase, the policy updates its expectations of long-term production potential. Once uncertainty has sufficiently collapsed as seen in time step 15 with hard rock mine \#2, and the expected value of the larger deposits becomes clearer, POMCPOW sequentially opens the hard-rock mines. Although these mines involve higher capital expenditure and higher emissions per tonne, their substantially larger reserves and lower operating cost justify the investment over the remaining horizon.

This staged expansion reflects belief-aware planning: initial low-risk production, systematic uncertainty reduction, and subsequent scale-up only when the expected discounted reward outweighs additional capital and environmental costs. In contrast, the heuristic policies lack this structured coupling between learning and sequential decision making. As shown in \cref{fig:action_timeline}, ExploreOnly continues exploring throughout the horizon, reducing uncertainty but generating negative profit and zero met demand. OneMineProfitMaximization opens only a single mine, selecting the larger hard-rock deposit to satisfy growing demand, but fails to adapt dynamically. DynamicProfitMaximizer opens a mine at each time step to maximize short-term profit, resulting in inefficient capital deployment and a lower percentage of met demand than POMCPOW. By synchronizing belief refinement with capital commitment, POMCPOW enables better-timed extraction decisions. This structured integration of learning and action selection yields higher discounted rewards and allows the policy to outperform the heuristics across profit, emissions, and demand-fulfillment metrics.

To enable a clear comparison across the different planning pricing models, we focus our analysis on the results at $\alpha = 0.5$. The complete evaluation data for various $\alpha$ values (0, 0.25, 0.5, 0.75, and 1.0) are provided in \cref{appendix:static} to \cref{appendix:Historic}.

\input{tables/0.5_Static}
Under static pricing, where market conditions remain nearly constant and future price signals provide little incentive for long-term planning, the OSL policy achieves a slightly higher discounted return, about 2.95\% above POMCPOW. Because production simultaneously generates revenue and yields precise geological information, short-term rollouts capture most of the attainable value without requiring deeper foresight when prices remain stable over time. However, POMCPOW still surpasses OSL across several operational metrics: it meets a higher share of demand (91.36\% vs.~89.18\%), achieves greater capacity utilization than OSL (0.64 vs.~0.63), and operates at a lower cost per tonne of lithium (\$8,036 vs.~\$8,089). These results suggest that while OSL’s myopic optimization performs well under stable market conditions by maximizing near-term returns since it plans one step ahead only, POMCPOW’s belief-based long term planning leads to more balanced, efficient operations; allocating capacity more effectively and reducing per-tonne costs through better sequential decision making over time.

\input{tables/0.5_linear}
As price rises steadily, myopic policies tend to open early to ride the slope, sometimes saturating the demand cap and paying OPEX on tonnes of lithium that do not sell. POMCPOW anticipates this and avoids over-opening, which is why it posts a higher discounted reward than profit-only heuristics while it still remains close on profit to the heuristics.

\input{tables/0.5_exponential}
Planning under exponential prices gives POMCPOW strong signals about when to wait and when to invest. It learns to delay opening until future prices justify the cost and then increases production when the returns are highest. When this policy is evaluated on historical prices, these timing decisions still pay off, which is why this setting produces POMCPOW’s highest discounted reward compared with the other price models. The heuristic policies do not learn this structure and often invest too early, which lowers their overall reward.

\input{tables/0.5_GBM}

Under GBM pricing, the combination of drift and volatility creates a much noisier planning environment than the static, linear, or exponential cases. When POMCPOW is planned under this setting and then evaluated on historical prices, it still achieves the highest discounted reward among the policies and meets the most demand, but its emissions are higher than in the other pricing models. This contrasts with the exponential and linear settings, where POMCPOW can time investment more clearly and keep emissions lower by concentrating production in the most profitable years. In the GBM case, the uncertain price path encourages a more reactive strategy that captures value effectively but results in greater operational intensity. Even so, POMCPOW remains more reliable than the heuristics in this volatile environment, showing that it handles uncertainty better than any fixed rule.

\input{tables/0.5_historical}
Even when planning and evaluation both use the historical price series, the discounted reward is not the highest of all the pricing models. Historical prices have a couple outliers, unlike the linear and exponential paths that rise more smoothly and give clearer future signals as seen in \cref{fig:pricing_models}. In those smoother environments, POMCPOW can time decisions more confidently and earn higher rewards. With historical prices, even though heuristic policies end up benefiting from early production that happens to align with the real price spikes, POMCPOW still performs strongly as it plans more optimally under uncertainty. 

Across all five pricing-model settings, POMCPOW consistently demonstrates stronger decision quality than the heuristic policies, achieving a higher discounted reward in almost every planning scenario. This shows that a belief-based planner is better equipped to handle uncertain price environments than human-like rule-based strategies, which are represented here by the heuristic baselines. Among these baselines, the OSL policy performs markedly better than the simple heuristics because, at each period, it enumerates all feasible joint actions and evaluates them using Monte Carlo rollouts under the $\alpha$ weighted profit–emissions reward, which is  similar to what is done in industry. This allows it to anticipate long-run production, depletion, and emission consequences of each decision, rather than relying solely on static lifetime formulas. However, unlike POMCPOW, it does not build a deeper belief tree or branch over future observation sequences. As a result, POMCPOW still achieves the highest discounted reward and demand satisfaction overall, while the OSL policy typically emerges as the second-best performer across our experiments.  One key feature of the environment that also makes POMCPOW forward looking search to outperform the heuristics is that the production system constrains the amount of sellable LCE to the company’s demand at each time step. Heuristic strategies that “open early and big” often commit capital and operating expenditures regardless of market absorption, later encountering demand ceilings that erode profitability. In contrast, POMCPOW anticipates these constraints, staging mine openings in alignment with evolving demand expectations. By doing so, it preserves margins while maintaining flexibility, mirroring the type of adaptive capacity planning rarely achievable through static heuristics.

Under static and linear pricing, POMCPOW performs comparably to One-Step Lookahead, since these settings contain weak long-term price signals that allow myopic strategies to capture much of the available value by producing early. Under exponential pricing, POMCPOW achieves its strongest discounted reward because the steady growth in prices provides clear incentives to wait, invest at the right time, and shift production toward the most profitable years. The GBM setting introduces drift and volatility, which makes future prices uncertain, yet POMCPOW still earns the highest discounted reward and meets the most demand among all policies. When planning is based on historical prices and then evaluated on the same data, POMCPOW continues to perform well, although the discounted reward does not reach the level seen in the exponential case because the historical series does not offer the same strong or predictable growth pattern. Across all tables, the POMCPOW Depth 1 variant remains at zero because a one-step tree cannot observe any future benefits. It only sees the immediate CAPEX and operating costs, with no long-term revenue or information gain to offset them. As a result, the optimal action from this limited view is always to avoid opening the mine, which leads to zero production, zero emissions, and a discounted reward of zero. 

When we expand these $\alpha$ values to 0.0, 0.25, 0.50, 0.75 and 1.0 the emission and profit scale according to the $\alpha$ values for all pricing model types which is seen in the Appendix below, but this can also be seen through the Pareto curve. The Pareto curve in \cref{fig:pareto_curve} illustrates how POMCPOW effectively captures the trade-off between profitability and emissions performance. A key advantage of this framework is its tunable $\alpha$-weight parameter, which allows decision-makers to explicitly define their priorities along the profit–emission spectrum. By adjusting $\alpha$, one can emphasize either economic returns (higher $\alpha$) or environmental responsibility (lower $\alpha$), providing a flexible decision space that aligns with different policy, corporate, or regulatory objectives. This tunability makes POMCPOW not only a multi objective optimization tool but also a practical planning instrument for real-world operations, where strategic goals may evolve over time or differ between stakeholders.

\begin{figure}[H]
    \centering   
    \input{figures/pareto_curve_linear_pricing}
    \caption{Pareto Curve for Linear Pricing}
    \label{fig:pareto_curve}
\end{figure}

The model’s performance across $\alpha$ values shows that POMCPOW adapts smoothly to shifting objective weights, maintaining stable and consistent trade-offs even under complex, nonlinear price dynamics. Heuristic policies, by contrast, exhibit rigid behavior optimizing for one goal at the expense of the other resulting in static profit and emissions given the $\alpha$ value. POMCPOW achieves superior balance because its planning process explicitly reasons about uncertainty: it evaluates the expected long-term reward of each action under multiple possible futures, rather than greedily maximizing short-term outcomes. This adaptability underscores POMCPOW’s strength as a robust, decision-aware planner capable of producing strategies tailored to the economic, environmental, and temporal priorities of lithium producers and policymakers alike. This can be further seen in \cref{appendix:static} to \cref{appendix:Historic} where POMCPOW almost always has the highest discounted reward for all $\alpha$ values (0, 0.25, 0.5, 0.75, and 1.0) across all the pricing models. Consequently, when $\alpha$ is tuned toward emissions minimization, the policy naturally delays or limits mine openings, while for profit-oriented settings, it identifies efficient extraction sequences that maximize returns without severely compromising sustainability. 
This dynamic behavior produces  interpretable movement along the empirical Pareto frontier as $\alpha$ varies, reflecting the planner’s ability to balance profitability and environmental performance under uncertainty.


Future work could extend the framework along several dimensions to better capture the strategic and regulatory environment of lithium production. One promising direction is the introduction of market feedback mechanisms, allowing production decisions to influence prices and competitor behavior rather than assuming prices evolve independently of firm actions. Endogenizing prices in this way would enable the model to account for how a producer’s output shapes market supply, demand, and ultimately price trajectories. Another important extension is to relax the assumption that mines operate at full capacity once opened; allowing production to adjust dynamically in response to demand, inventories, or price conditions would provide a more flexible and realistic operational model. A multi agent formulation, where multiple mining companies interact, compete, or potentially cooperate, would further enrich the analysis by capturing market power and strategic behavior among producers. Finally, incorporating global or national regulatory constraints, such as carbon pricing schemes or extraction quotas per deposit, would allow the framework to evaluate how policy interventions shape optimal production and investment strategies.

\section{Conclusions}
This work casts lithium production planning as a sequential, information-aware decision problem and introduces a POMDP framework that jointly models geological uncertainty, demand, price dynamics, and technology choice between DLE and hard-rock mining. Using the online planner POMCPOW, we show that belief-aware strategies generally outperform a suite of human-interpretable heuristics across static, linear, exponential, GBM price regimes and even historical, delivering the highest (or near-highest) discounted reward while maintaining competitive emissions and strong demand satisfaction. A key insight is that when and where uncertainty is reduced creates value: by updating beliefs from exploration and production signals, the planner avoids “open-early and big” pitfalls that collide with demand ceilings, instead sequencing exploration, opening, and restoration to align capacity with market absorption and to sell into advantageous years. The tunable $\alpha$ parameter traces a smooth empirical Pareto frontier between profit and emissions, enabling decision-makers to encode priorities without making an suboptimal decision. This implies companies should wait to make big, irreversible investments until they have more confidence in the data or until future prices and profits clearly justify expansion. Exploration then focuses on the areas where new information would be most valuable for reducing uncertainty. Technology choices should also depend on the situation: favoring lower-cost and lower-emission options when demand or prices are uncertain, and switching to larger, higher-output operations when profits and remaining resources make it worthwhile. By putting a clear cost on emissions, the framework helps connect financial decisions with sustainability goals. Limitations, such as exogenous prices without market feedback, simplified demand representation, streamlined techno-environmental modules, point to extensions. Nonetheless, framing lithium production decision making as a POMDP yields adaptive, transparent, and tunable strategies that help investors, operators, and policymakers stage the right capacity at the right time with clearer visibility into both economic and environmental outcomes.

\section{Acknowledgments}
The authors wish to thank the Mineral X team at Stanford, including Greg Forbes, and industry professionals such as Shane Tragethon for their thoughtful discussions and expertise on lithium industry practices.

\section{Appendix}

In this appendix, we summarize the data used to parameterize the model, including cost estimates, deposit sizes, and other key inputs, as well as the historical production and price data for lithium \cref{appendix:historic_price}. The full results for all $\alpha$ values (0, 0.25, 0.5, 0.75, and 1.0) are presented in the following sections: the Static Pricing Model Results (\cref{appendix:static}), Linear Pricing Model Results (\cref{appendix:linear}), Exponential Pricing Model Results (\cref{appendix:exponential}), Geometric Brownian Motion Pricing Model Results (\cref{appendix:gbm}), and Historic Pricing Model Results (\cref{appendix:Historic}).

\appendix

\section{Deposit Cost and Size Comparison: DLE vs. Hard Rock}
\label{appendix:dle_vs_hardrock}
\input{tables/costs_deposit_size_DLE_vs_hard_rock}

\section{Model Parameters}
\label{appendix:params}
\input{tables/model_params}

\section{Historic Lithium Data}
\label{appendix:historic_price}
\input{tables/historic_data}

\section{Observation Noise}
\label{appendix:observation_noise}
\input{tables/observation_noise}

\section{Price Equation Constants}
\label{appendix:price_equation_constants}
\input{tables/price_equation_constants}

\section{Static Pricing Model Results}
\label{appendix:static}
\input{tables/all_alphas_static}

\section{Linear Pricing Model Results}
\label{appendix:linear}
\input{tables/all_alphas_linear}

\section{Exponential Pricing Model Results}
\label{appendix:exponential}
\input{tables/all_alphas_exponential}

\section{Geometric Brownian Motion Pricing Model Results}
\label{appendix:gbm}
\input{tables/all_alphas_GBM}

\section{Historic Pricing Model Results}
\label{appendix:Historic}
\input{tables/all_alphas_historical}

\printbibliography

\end{document}

%% file: figures/pricing_models_graph.tex
\begin{tikzpicture}
  \begin{axis}[
    title={Lithium Price Models Over Time},
    title style={font=\small, yshift=1ex},
    width=14cm, height=9cm,
    xmin=1995, xmax=2024,
    ymin=0, ymax=80000,
    xlabel={Year}, ylabel={Price (USD/LCE metric ton)},
    grid=both, grid style={opacity=0.2},
    tick align=outside,
    every axis plot/.append style={line width=1pt},
    scaled y ticks=false,
    scaled x ticks=false,
    xticklabel style={
      /pgf/number format/.cd,
      1000 sep={}
    },
    yticklabel style={
      /pgf/number format/.cd,
      fixed,
      fixed zerofill,
      precision=0,
      1000 sep={,} 
    },
    legend style={
      font=\small,
      draw=black, thick,
      fill=white, fill opacity=0.9, text opacity=1,
      rounded corners=2pt,
      at={(0.02,0.98)}, anchor=north west,
      legend cell align=left
    },
  ]

  \pgfmathsetmacro{\startyear}{1995}
  \pgfmathsetmacro{\T}{29}
  \pgfmathsetmacro{\pzerostatic}{10355.33}
  \pgfmathsetmacro{\pzerolinear}{2381.38}
  \pgfmathsetmacro{\pzeroexp}{917.36}
  \pgfmathsetmacro{\pzerogbm}{10000.0}
  \pgfmathsetmacro{\alphaslope}{878.39}
  \pgfmathsetmacro{\lambdaexp}{0.1261}
  \pgfmathsetmacro{\mudrift}{0.033}   
  \pgfmathsetmacro{\sigmavol}{0.23}  
  \pgfmathsetmacro{\staticNoise}{0.05}
  \pgfmathsetmacro{\linearNoise}{0.10}
  \pgfmathsetmacro{\expNoise}{0.10}

  \addplot[
    name path=StaticMean,
    domain=0:\T, samples=\numexpr \T+1 \relax,
    color=gray!60, very thick, solid,
    legend image post style={color=gray!60, very thick, solid}
  ] ({\startyear + x},{\pzerostatic});
  \addlegendentry{Static}
  \addplot[name path=StaticUpper, draw=none, domain=0:\T, samples=\numexpr \T+1 \relax, forget plot]
    ({\startyear + x},{\pzerostatic*(1+\staticNoise)});
  \addplot[name path=StaticLower, draw=none, domain=0:\T, samples=\numexpr \T+1 \relax, forget plot]
    ({\startyear + x},{max(\pzerostatic*(1-\staticNoise),1000)});
  \addplot[fill=gray!15, draw=none, opacity=0.3, forget plot]
    fill between[of=StaticUpper and StaticLower];

  \addplot[
    name path=LinearMean,
    domain=0:\T, samples=\numexpr \T+1 \relax,
    color=cyan!70!blue, thick, dashed,
    legend image post style={color=cyan!70!blue, thick, dashed}
  ] ({\startyear + x},{\pzerolinear + \alphaslope*x});
  \addlegendentry{Linear}
  \addplot[name path=LinearUpper, draw=none, domain=0:\T, samples=\numexpr \T+1 \relax, forget plot]
    ({\startyear + x},{(\pzerolinear + \alphaslope*x)*(1+\linearNoise)});
  \addplot[name path=LinearLower, draw=none, domain=0:\T, samples=\numexpr \T+1 \relax, forget plot]
    ({\startyear + x},{max((\pzerolinear + \alphaslope*x)*(1-\linearNoise),1000)});
  \addplot[fill=cyan!20, draw=none, opacity=0.25, forget plot]
    fill between[of=LinearUpper and LinearLower];

  \addplot[
    name path=ExpMean,
    domain=0:\T, samples=\numexpr \T+1 \relax,
    color=black!20!green, thick, dash dot,
    legend image post style={color=black!20!green, thick, dash dot}
  ] ({\startyear + x},{\pzeroexp*exp(\lambdaexp*x)});
  \addlegendentry{Exponential}
  \addplot[name path=ExpUpper, draw=none, domain=0:\T, samples=\numexpr \T+1 \relax, forget plot]
    ({\startyear + x},{\pzeroexp*exp(\lambdaexp*x)*(1+\expNoise)});
  \addplot[name path=ExpLower, draw=none, domain=0:\T, samples=\numexpr \T+1 \relax, forget plot]
    ({\startyear + x},{max(\pzeroexp*exp(\lambdaexp*x)*(1-\expNoise),1000)});
  \addplot[fill=lime!25, draw=none, opacity=0.25, forget plot]
    fill between[of=ExpUpper and ExpLower];

  \addplot[
    name path=GBMMean,
    domain=0:\T, samples=\numexpr \T+1 \relax,
    color=orange!90!yellow, ultra thick, dotted,
    mark=*, mark size=2pt,
    mark options={fill=orange!80!yellow},
    legend image post style={color=orange!90!yellow, ultra thick, dotted, mark=*, mark size=2pt}
  ] ({\startyear + x},{\pzerogbm*exp(\mudrift*x)});
  \addlegendentry{GBM (expected)}
  \addplot[name path=GBMUpper, draw=none, domain=0:\T, samples=\numexpr \T+1 \relax, forget plot]
    ({\startyear + x},{\pzerogbm*exp(\mudrift*x) + \pzerogbm*exp(\mudrift*x)*sqrt(exp((\sigmavol*\sigmavol)*x)-1)});
  \addplot[name path=GBMLower, draw=none, domain=0:\T, samples=\numexpr \T+1 \relax, forget plot]
    ({\startyear + x},{max(\pzerogbm*exp(\mudrift*x) - \pzerogbm*exp(\mudrift*x)*sqrt(exp((\sigmavol*\sigmavol)*x)-1),1000)});
  \addplot[fill=orange!30, draw=none, opacity=0.25, forget plot]
    fill between[of=GBMUpper and GBMLower];

  \addplot+[
    mark=*,
    mark size=1.5pt,
    color=pink!80!red, very thick, solid,
    legend image post style={color=pink!80!red, very thick, solid, mark=*}
  ] coordinates {
      (1995,3200) (1996,3400) (1997,3800) (1998,4000) (1999,4200)
      (2000,4400) (2001,4600) (2002,4800) (2003,5000) (2004,5200)
      (2005,5400) (2006,5600) (2007,6000) (2008,6400) (2009,5800)
      (2010,5180) (2011,5180) (2012,6060) (2013,6800) (2014,6690)
      (2015,6500) (2016,8650) (2017,15000) (2018,16000) (2019,12100)
      (2020,10100) (2021,14200) (2022,71100) (2023,41300) (2024,14000)
  };
  \addlegendentry{Historical}
  \path[name path=HistUpper] plot coordinates {
      (1995,3520) (1996,3740) (1997,4180) (1998,4400) (1999,4620)
      (2000,4840) (2001,5060) (2002,5280) (2003,5500) (2004,5720)
      (2005,5940) (2006,6160) (2007,6600) (2008,7040) (2009,6380)
      (2010,5698) (2011,5698) (2012,6666) (2013,7480) (2014,7359)
      (2015,7150) (2016,9515) (2017,16500) (2018,17600) (2019,13310)
      (2020,11110) (2021,15620) (2022,78210) (2023,45430) (2024,15400)
  };
  \path[name path=HistLower] plot coordinates {
      (1995,2880) (1996,3060) (1997,3420) (1998,3600) (1999,3780)
      (2000,3960) (2001,4140) (2002,4320) (2003,4500) (2004,4680)
      (2005,4860) (2006,5040) (2007,5400) (2008,5760) (2009,5220)
      (2010,4662) (2011,4662) (2012,5454) (2013,6120) (2014,6021)
      (2015,5850) (2016,7785) (2017,13500) (2018,14400) (2019,10890)
      (2020,9090) (2021,12780) (2022,63990) (2023,37170) (2024,12600)
  };
  \addplot[fill=pink!35, draw=none, opacity=50, forget plot]
    fill between[of=HistUpper and HistLower];

  \end{axis}
\end{tikzpicture}

%% file: figures/belief_uncertainty_grid.tex

\begin{figure}[H]
\centering

\pgfplotsset{
  every axis/.append style={
    tick label style={font=\scriptsize},
    label style={font=\small},
    title style={font=\small, yshift=3pt},
    xlabel near ticks,
    ylabel near ticks,
    xlabel={Time Step},
    ylabel={Belief Uncertainty (tonnes Li)},
    xtick align=outside,
    ytick align=outside,
    grid=major,
    grid style={line width=.2pt, draw=gray!20},
    enlarge y limits=0.03,
    scaled y ticks=false,
    yticklabel style={/pgf/number format/fixed}, 
  },
  pomstyle/.style={lightblue,  thick,  mark=*,        mark size=1.2pt, error bars/.cd, y dir=both, y explicit},
  profstyle/.style={lightred,   thick,  dashed,        mark=square*,   mark size=1.2pt, error bars/.cd, y dir=both, y explicit},
  explstyle/.style={lightgreen, thick,  dotted,        mark=triangle*, mark size=1.4pt, error bars/.cd, y dir=both, y explicit}
}

\begin{tikzpicture}
\begin{groupplot}[
  group style={
    group size=2 by 2,
    horizontal sep=14mm,         
    vertical sep=12mm,           
    x descriptions at=edge bottom,
    y descriptions at=edge left   
  },
  width=0.46\textwidth,          
  height=5.2cm,
  xmin=0, xmax=29
]

\nextgroupplot[title={Site 1 (DLE)}, legend to name=depositLegend, legend columns=3, legend style={/tikz/every even column/.append style={column sep=7pt}, draw=none, fill=none, font=\scriptsize}]
\addplot[pomstyle] coordinates {
    (0, 39000.0) +- (0, 0.0) (1, 37038.07) +- (0, 1658.93) (2, 36139.06) +- (0, 2342.24) (3, 34343.54) +- (0, 2246.13) (4, 27377.33) +- (0, 11356.24) (5, 23202.17) +- (0, 13581.13) (6, 18500.25) +- (0, 13812.25) (7, 15681.97) +- (0, 13255.12) (8, 14840.92) +- (0, 12996.56) (9, 13094.06) +- (0, 12740.62) (10, 12670.67) +- (0, 12722.54) (11, 8130.37) +- (0, 10148.73) (12, 6200.66) +- (0, 8610.85) (13, 5775.15) +- (0, 8519.77) (14, 5561.73) +- (0, 8591.69) (15, 2636.07) +- (0, 1370.75) (16, 2069.45) +- (0, 1163.19) (17, 1817.88) +- (0, 1001.29) (18, 1740.96) +- (0, 790.57) (19, 1413.32) +- (0, 887.33) (20, 1233.36) +- (0, 815.11) (21, 1258.37) +- (0, 600.22) (22, 1020.38) +- (0, 633.27) (23, 865.22) +- (0, 616.88) (24, 824.43) +- (0, 577.19) (25, 676.26) +- (0, 514.41) (26, 891.36) +- (0, 391.32) (27, 1061.3) +- (0, 136.23) (28, 1139.0) +- (0, 79.42) (29, 1168.31) +- (0, 52.55)
};
\addplot[profstyle] coordinates {
    (0, 39000.0) +- (0, 0.0) (1, 39000.0) +- (0, 0.0) (2, 39000.0) +- (0, 0.0) (3, 39000.0) +- (0, 0.0) (4, 39000.0) +- (0, 0.0) (5, 39000.0) +- (0, 0.0) (6, 39000.0) +- (0, 0.0) (7, 39000.0) +- (0, 0.0) (8, 39000.0) +- (0, 0.0) (9, 39000.0) +- (0, 0.0) (10, 39000.0) +- (0, 0.0) (11, 39000.0) +- (0, 0.0) (12, 39000.0) +- (0, 0.0) (13, 39000.0) +- (0, 0.0) (14, 39000.0) +- (0, 0.0) (15, 39000.0) +- (0, 0.0) (16, 39000.0) +- (0, 0.0) (17, 39000.0) +- (0, 0.0) (18, 39000.0) +- (0, 0.0) (19, 39000.0) +- (0, 0.0) (20, 39000.0) +- (0, 0.0) (21, 39000.0) +- (0, 0.0) (22, 39000.0) +- (0, 0.0) (23, 39000.0) +- (0, 0.0) (24, 39000.0) +- (0, 0.0) (25, 39000.0) +- (0, 0.0) (26, 39000.0) +- (0, 0.0) (27, 39000.0) +- (0, 0.0) (28, 39000.0) +- (0, 0.0) (29, 39000.0) +- (0, 0.0)
};
\addplot[explstyle] coordinates {
    (0, 39000.0) +- (0, 0.0) (1, 35702.85) +- (0, 78.26) (2, 33128.31) +- (0, 108.02) (3, 31055.96) +- (0, 129.02) (4, 29335.32) +- (0, 126.3) (5, 27876.3) +- (0, 116.25) (6, 26623.8) +- (0, 103.1) (7, 25526.11) +- (0, 95.04) (8, 24555.38) +- (0, 86.45) (9, 23689.08) +- (0, 77.81) (10, 22908.49) +- (0, 70.72) (11, 22200.44) +- (0, 64.38) (12, 21554.22) +- (0, 58.93) (13, 20961.24) +- (0, 54.2) (14, 20414.59) +- (0, 50.08) (15, 19908.57) +- (0, 46.48) (16, 19438.47) +- (0, 43.3) (17, 19000.09) +- (0, 40.5) (18, 18589.98) +- (0, 38.06) (19, 18205.2) +- (0, 35.99) (20, 17843.47) +- (0, 34.12) (21, 17502.75) +- (0, 32.23) (22, 17180.8) +- (0, 30.53) (23, 16876.03) +- (0, 28.96) (24, 16586.93) +- (0, 27.5) (25, 16312.2) +- (0, 26.16) (26, 16050.69) +- (0, 24.93) (27, 15801.37) +- (0, 23.79) (28, 15563.32) +- (0, 22.73) (29, 15335.71) +- (0, 21.75)
};
\legend{POMCPOW, Profit Maximizer, Explore Only}

\nextgroupplot[title={Site 2 (DLE)}, legend=false]
\addplot[pomstyle] coordinates {
    (0, 39000.0) +- (0, 0.0) (1, 37458.19) +- (0, 1706.95) (2, 35646.71) +- (0, 2105.19) (3, 34271.28) +- (0, 2373.25) (4, 31073.35) +- (0, 7364.04) (5, 28227.52) +- (0, 9597.84) (6, 24322.18) +- (0, 12277.7) (7, 23289.04) +- (0, 12331.56) (8, 20824.89) +- (0, 12694.23) (9, 16040.19) +- (0, 13323.46) (10, 15147.65) +- (0, 13221.86) (11, 12907.97) +- (0, 12455.99) (12, 11239.13) +- (0, 12124.11) (13, 8146.14) +- (0, 10434.34) (14, 6025.78) +- (0, 8295.91) (15, 4437.56) +- (0, 6555.18) (16, 4122.07) +- (0, 6587.25) (17, 3840.67) +- (0, 6306.42) (18, 3622.04) +- (0, 6058.84) (19, 2198.06) +- (0, 1065.91) (20, 1992.96) +- (0, 674.27) (21, 1768.53) +- (0, 690.89) (22, 1724.03) +- (0, 468.16) (23, 1592.93) +- (0, 546.74) (24, 1245.68) +- (0, 729.7) (25, 1369.83) +- (0, 409.45) (26, 1142.4) +- (0, 643.74) (27, 1030.46) +- (0, 620.38) (28, 801.17) +- (0, 648.72) (29, 821.35) +- (0, 502.31)
};
\addplot[profstyle] coordinates {
    (0, 39000.0) +- (0, 0.0) (1, 39000.0) +- (0, 0.0) (2, 39000.0) +- (0, 0.0) (3, 39000.0) +- (0, 0.0) (4, 39000.0) +- (0, 0.0) (5, 39000.0) +- (0, 0.0) (6, 39000.0) +- (0, 0.0) (7, 39000.0) +- (0, 0.0) (8, 39000.0) +- (0, 0.0) (9, 39000.0) +- (0, 0.0) (10, 39000.0) +- (0, 0.0) (11, 39000.0) +- (0, 0.0) (12, 39000.0) +- (0, 0.0) (13, 39000.0) +- (0, 0.0) (14, 39000.0) +- (0, 0.0) (15, 39000.0) +- (0, 0.0) (16, 39000.0) +- (0, 0.0) (17, 39000.0) +- (0, 0.0) (18, 39000.0) +- (0, 0.0) (19, 39000.0) +- (0, 0.0) (20, 39000.0) +- (0, 0.0) (21, 39000.0) +- (0, 0.0) (22, 39000.0) +- (0, 0.0) (23, 39000.0) +- (0, 0.0) (24, 39000.0) +- (0, 0.0) (25, 39000.0) +- (0, 0.0) (26, 39000.0) +- (0, 0.0) (27, 39000.0) +- (0, 0.0) (28, 39000.0) +- (0, 0.0) (29, 39000.0) +- (0, 0.0)
};
\addplot[explstyle] coordinates {
    (0, 39000.0) +- (0, 0.0) (1, 35719.8) +- (0, 80.39) (2, 33166.98) +- (0, 131.75) (3, 31095.76) +- (0, 154.31) (4, 29375.1) +- (0, 165.77) (5, 27914.19) +- (0, 167.28) (6, 26646.78) +- (0, 180.99) (7, 25532.28) +- (0, 205.48) (8, 24548.74) +- (0, 227.98) (9, 23668.19) +- (0, 257.21) (10, 22883.01) +- (0, 255.83) (11, 22173.78) +- (0, 245.68) (12, 21525.61) +- (0, 240.64) (13, 20930.58) +- (0, 238.06) (14, 20385.02) +- (0, 225.43) (15, 19879.58) +- (0, 215.85) (16, 19409.47) +- (0, 209.1) (17, 18972.94) +- (0, 196.46) (18, 18564.77) +- (0, 184.49) (19, 18181.88) +- (0, 173.64) (20, 17821.76) +- (0, 163.81) (21, 17482.22) +- (0, 154.87) (22, 17161.39) +- (0, 146.72) (23, 16857.6) +- (0, 139.25) (24, 16569.39) +- (0, 132.4) (25, 16295.48) +- (0, 126.09) (26, 16034.73) +- (0, 120.27) (27, 15786.1) +- (0, 114.89) (28, 15548.7) +- (0, 109.89) (29, 15321.69) +- (0, 105.25)
};

\nextgroupplot[title={Site 3 (Hard Rock)}, legend=false]
\addplot[pomstyle] coordinates {
    (0, 173756.0) +- (0, 0.0) (1, 154431.27) +- (0, 18711.09) (2, 142078.02) +- (0, 17812.48) (3, 132048.47) +- (0, 20250.59) (4, 126238.9) +- (0, 22248.84) (5, 117404.55) +- (0, 20988.49) (6, 114137.61) +- (0, 21793.85) (7, 94693.14) +- (0, 35710.41) (8, 72677.1) +- (0, 45785.45) (9, 55531.3) +- (0, 44246.75) (10, 53060.98) +- (0, 44129.36) (11, 52169.4) +- (0, 43979.84) (12, 51012.54) +- (0, 42529.56) (13, 50240.6) +- (0, 42087.7) (14, 47739.17) +- (0, 39302.23) (15, 46666.35) +- (0, 38117.89) (16, 45262.02) +- (0, 36633.99) (17, 45085.33) +- (0, 36533.88) (18, 43744.34) +- (0, 35278.74) (19, 42581.37) +- (0, 33655.04) (20, 41138.73) +- (0, 32398.03) (21, 36818.51) +- (0, 30972.99) (22, 32613.06) +- (0, 29976.61) (23, 31858.77) +- (0, 29848.32) (24, 27649.04) +- (0, 27480.19) (25, 26798.14) +- (0, 26670.28) (26, 10648.86) +- (0, 2871.28) (27, 9275.68) +- (0, 860.84) (28, 8687.8) +- (0, 0.0) (29, 8687.8) +- (0, 0.0)
};
\addplot[profstyle] coordinates {
    (0, 173756.0) +- (0, 0.0) (1, 14944.42) +- (0, 0.0) (2, 10586.9) +- (0, 0.0) (3, 8687.8) +- (0, 0.0) (4, 8687.8) +- (0, 0.0) (5, 8687.8) +- (0, 0.0) (6, 8687.8) +- (0, 0.0) (7, 8687.8) +- (0, 0.0) (8, 8687.8) +- (0, 0.0) (9, 8687.8) +- (0, 0.0) (10, 8687.8) +- (0, 0.0) (11, 8687.8) +- (0, 0.0) (12, 8687.8) +- (0, 0.0) (13, 8687.8) +- (0, 0.0) (14, 8687.8) +- (0, 0.0) (15, 8687.8) +- (0, 0.0) (16, 8687.8) +- (0, 0.0) (17, 8687.8) +- (0, 0.0) (18, 8687.8) +- (0, 0.0) (19, 8687.8) +- (0, 0.0) (20, 8687.8) +- (0, 0.0) (21, 8687.8) +- (0, 0.0) (22, 8687.8) +- (0, 0.0) (23, 8687.8) +- (0, 0.0) (24, 8687.8) +- (0, 0.0) (25, 8687.8) +- (0, 0.0) (26, 8687.8) +- (0, 0.0) (27, 8687.8) +- (0, 0.0) (28, 8687.8) +- (0, 0.0) (29, 8687.8) +- (0, 0.0)
};
\addplot[explstyle] coordinates {
    (0, 173756.0) +- (0, 0.0) (1, 137521.25) +- (0, 2.4) (2, 117339.44) +- (0, 1.52) (3, 104041.2) +- (0, 1.06) (4, 94434.01) +- (0, 0.79) (5, 87075.58) +- (0, 0.62) (6, 81206.49) +- (0, 0.5) (7, 76383.8) +- (0, 0.42) (8, 72329.47) +- (0, 0.36) (9, 68859.01) +- (0, 0.31) (10, 65844.5) +- (0, 0.27) (11, 63194.12) +- (0, 0.24) (12, 60840.02) +- (0, 0.21) (13, 58730.8) +- (0, 0.19) (14, 56826.77) +- (0, 0.17) (15, 55096.67) +- (0, 0.16) (16, 53515.55) +- (0, 0.14) (17, 52063.17) +- (0, 0.13) (18, 50722.96) +- (0, 0.12) (19, 49481.2) +- (0, 0.11) (20, 48326.39) +- (0, 0.11) (21, 47248.83) +- (0, 0.1) (22, 46240.28) +- (0, 0.09) (23, 45293.68) +- (0, 0.09) (24, 44402.92) +- (0, 0.08) (25, 43562.74) +- (0, 0.08) (26, 42768.51) +- (0, 0.07) (27, 42016.2) +- (0, 0.07) (28, 41302.23) +- (0, 0.07) (29, 40623.47) +- (0, 0.06)
};

\nextgroupplot[title={Site 4 (Hard Rock)}, legend=false]
\addplot[pomstyle] coordinates {
    (0, 173756.0) +- (0, 0.0) (1, 159262.51) +- (0, 18373.85) (2, 149049.43) +- (0, 25108.82) (3, 139570.65) +- (0, 23813.02) (4, 129176.01) +- (0, 22364.37) (5, 116747.61) +- (0, 34223.55) (6, 104133.38) +- (0, 42648.61) (7, 81301.82) +- (0, 45928.52) (8, 60529.0) +- (0, 48966.94) (9, 58073.68) +- (0, 48102.09) (10, 56419.71) +- (0, 47292.57) (11, 52853.96) +- (0, 43454.17) (12, 51972.16) +- (0, 42677.43) (13, 51010.16) +- (0, 41677.58) (14, 49377.45) +- (0, 40036.99) (15, 45041.99) +- (0, 39847.3) (16, 42608.84) +- (0, 37501.64) (17, 42167.65) +- (0, 37278.71) (18, 41575.84) +- (0, 36680.35) (19, 40984.04) +- (0, 36061.66) (20, 39537.79) +- (0, 34425.52) (21, 39066.53) +- (0, 33951.93) (22, 23765.16) +- (0, 27409.89) (23, 17896.37) +- (0, 21757.07) (24, 16956.46) +- (0, 21598.96) (25, 16677.55) +- (0, 21288.49) (26, 9481.61) +- (0, 2095.06) (27, 8926.41) +- (0, 629.79) (28, 8687.8) +- (0, 0.0) (29, 8687.8) +- (0, 0.0)
};
\addplot[profstyle] coordinates {
    (0, 173756.0) +- (0, 0.0) (1, 173756.0) +- (0, 0.0) (2, 173756.0) +- (0, 0.0) (3, 173756.0) +- (0, 0.0) (4, 173756.0) +- (0, 0.0) (5, 173756.0) +- (0, 0.0) (6, 173756.0) +- (0, 0.0) (7, 173756.0) +- (0, 0.0) (8, 173756.0) +- (0, 0.0) (9, 173756.0) +- (0, 0.0) (10, 173756.0) +- (0, 0.0) (11, 173756.0) +- (0, 0.0) (12, 173756.0) +- (0, 0.0) (13, 173756.0) +- (0, 0.0) (14, 173756.0) +- (0, 0.0) (15, 173756.0) +- (0, 0.0) (16, 173756.0) +- (0, 0.0) (17, 173756.0) +- (0, 0.0) (18, 173756.0) +- (0, 0.0) (19, 173756.0) +- (0, 0.0) (20, 173756.0) +- (0, 0.0) (21, 173756.0) +- (0, 0.0) (22, 173756.0) +- (0, 0.0) (23, 173756.0) +- (0, 0.0) (24, 173756.0) +- (0, 0.0) (25, 173756.0) +- (0, 0.0) (26, 173756.0) +- (0, 0.0) (27, 173756.0) +- (0, 0.0) (28, 173756.0) +- (0, 0.0) (29, 173756.0) +- (0, 0.0)
};
\addplot[explstyle] coordinates {
    (0, 173756.0) +- (0, 0.0) (1, 137522.18) +- (0, 0.44) (2, 117340.03) +- (0, 0.27) (3, 104041.61) +- (0, 0.19) (4, 94434.32) +- (0, 0.14) (5, 87075.83) +- (0, 0.11) (6, 81206.68) +- (0, 0.09) (7, 76383.96) +- (0, 0.08) (8, 72329.61) +- (0, 0.06) (9, 68859.13) +- (0, 0.06) (10, 65844.6) +- (0, 0.05) (11, 63194.21) +- (0, 0.04) (12, 60840.1) +- (0, 0.04) (13, 58730.88) +- (0, 0.03) (14, 56826.84) +- (0, 0.03) (15, 55096.73) +- (0, 0.03) (16, 53515.6) +- (0, 0.03) (17, 52063.22) +- (0, 0.02) (18, 50723.01) +- (0, 0.02) (19, 49481.24) +- (0, 0.02) (20, 48326.43) +- (0, 0.02) (21, 47248.87) +- (0, 0.02) (22, 46240.32) +- (0, 0.02) (23, 45293.71) +- (0, 0.02) (24, 44402.96) +- (0, 0.01) (25, 43562.77) +- (0, 0.01) (26, 42768.54) +- (0, 0.01) (27, 42016.22) +- (0, 0.01) (28, 41302.26) +- (0, 0.01) (29, 40623.5) +- (0, 0.01)
};

\end{groupplot}

\path (current bounding box.south);
\node at ([yshift=-0.7em]current bounding box.south) {\ref{depositLegend}};

\end{tikzpicture}

\caption{Average over 30 simulations of deposit belief uncertainty evolution for sites 1-4 under three policies (POMCPOW, Profit Maximizer, Explore Only).}
\label{fig:deposit_uncertainty_grid}
\end{figure}

%% file: tables/0.5_static.tex

\begin{table}[H]
\centering
\caption{Policy comparison planned with \textbf{static} prices, evaluated on historical prices ($\alpha=0.5$)}
\label{tab:p1e6_alpha05}
\renewcommand{\arraystretch}{1.1}
\resizebox{\textwidth}{!}{%
\begin{tabular}{@{}l r r r r@{}}
\toprule
Policy 
& Profit (B\$) $\uparrow$ 
& Emissions Cost (M\$) $\downarrow$ 
& Disc.\ Reward $\uparrow$ 
& Demand Met (\%) $\uparrow$ \\
\midrule
\textbf{POMCPOW} 
& $16.70 \pm 4.26$ 
& $0.94 \pm 0.06$ 
& $2.96\times10^{-1} \pm 9.11\times10^{-2}$ 
& $\mathbf{91.36}$ \\

OneStepLookahead 
& $\mathbf{17.00 \pm 4.21}$ 
& $0.93 \pm 0.08$ 
& $\mathbf{3.05\times10^{-1} \pm 8.90\times10^{-2}}$ 
& 89.18 \\

DynamicProfitMaximizer 
& $14.80 \pm 1.98$ 
& $0.91 \pm 0.03$ 
& $1.67\times10^{-1} \pm 4.27\times10^{-2}$ 
& 90.40 \\

RandomHeuristicPolicy 
& $14.10 \pm 1.76$ 
& $0.91 \pm 0.04$ 
& $1.55\times10^{-1} \pm 3.88\times10^{-2}$ 
& 89.93 \\

OneMineProfitMaximization 
& $9.14 \pm 0.69$ 
& $0.88 \pm 0.01$ 
& $1.28\times10^{-1} \pm 1.53\times10^{-2}$ 
& 69.28 \\

DynamicEmissionMinimizer 
& $1.87 \pm 2.81$ 
& $0.41 \pm 0.08$ 
& $1.46\times10^{-3} \pm 6.09\times10^{-2}$ 
& 49.44 \\

POMCPOW (Depth 1) 
& $0.00 \pm 0.00$ 
& $\boldsymbol{0.00 \pm 0.00}$ 
& $0.00 \pm 0.00$ 
& 0.00 \\

ExploreOnly 
& $-0.29 \pm 0.00$ 
& $\boldsymbol{0.00 \pm 0.00}$ 
& $-9.78\times10^{-3} \pm 1.75\times10^{-18}$ 
& 0.00 \\

OneMineEmissionMinimization 
& $-0.95 \pm 0.16$ 
& $0.23 \pm 0.06$ 
& $-5.53\times10^{-2} \pm 3.87\times10^{-3}$ 
& 27.17 \\
\bottomrule
\end{tabular}%
}
\end{table}

%% file: tables/0.5_linear.tex
\begin{table}[H]
\centering
\caption{Policy comparison planned with \textbf{linear} prices, evaluated on historical prices ($\alpha=0.5$)}
\label{tab:p2e6_alpha_0.5}
\renewcommand{\arraystretch}{1.1}
\resizebox{\textwidth}{!}{%
\begin{tabular}{@{}l r r r r@{}}
\toprule
Policy 
& Profit (B\$) $\uparrow$ 
& Emissions Cost (M\$) $\downarrow$ 
& Disc.\ Reward $\uparrow$ 
& Demand Met (\%) $\uparrow$ \\
\midrule
\textbf{POMCPOW} 
& $\boldsymbol{17.47 \pm 4.45}$ 
& $0.92 \pm 0.10$ 
& $\boldsymbol{3.04\times10^{-1} \pm 9.74\times10^{-2}}$ 
& 87.70 \\

OneStepLookahead 
& $15.80 \pm 3.83$ 
& $0.86 \pm 0.05$ 
& $2.54\times10^{-1} \pm 8.72\times10^{-2}$ 
& 86.17 \\

DynamicProfitMaximizer 
& $14.77 \pm 1.98$ 
& $0.91 \pm 0.03$ 
& $1.67\times10^{-1} \pm 4.27\times10^{-2}$ 
& \textbf{90.40} \\

RandomHeuristicPolicy 
& $14.06 \pm 1.76$ 
& $0.91 \pm 0.04$ 
& $1.55\times10^{-1} \pm 3.88\times10^{-2}$ 
& 89.93 \\

OneMineProfitMaximization 
& $9.14 \pm 0.69$ 
& $0.88 \pm 0.01$ 
& $1.28\times10^{-1} \pm 1.53\times10^{-2}$ 
& 69.28 \\

DynamicEmissionMinimizer 
& $1.87 \pm 2.81$ 
& $0.41 \pm 0.08$ 
& $1.46\times10^{-3} \pm 6.09\times10^{-2}$ 
& 49.44 \\

POMCPOW (Depth 1) 
& $0.00 \pm 0.00$ 
& $\boldsymbol{0.00 \pm 0.00}$ 
& $0.00 \pm 0.00$ 
& 0.00 \\

ExploreOnly 
& $-0.29 \pm 0.00$ 
& $\boldsymbol{0.00 \pm 0.00}$ 
& $-9.78\times10^{-3} \pm 1.75\times10^{-18}$ 
& 0.00 \\

OneMineEmissionMinimization 
& $-0.95 \pm 0.16$ 
& $0.23 \pm 0.06$ 
& $-5.53\times10^{-2} \pm 3.87\times10^{-3}$ 
& 27.17 \\
\bottomrule
\end{tabular}%
}
\end{table}

%% file: tables/0.5_exponential.tex

\begin{table}[H]
\centering
\caption{Policy comparison planned with \textbf{exponential} prices, evaluated on historical prices ($\alpha=0.5$)}
\label{tab:alpha05_results}
\resizebox{\textwidth}{!}{%
\begin{tabular}{@{}l r r r r@{}}
\toprule
Policy 
& Profit (B\$) $\uparrow$ 
& Emissions Cost (M\$) $\downarrow$ 
& Disc.\ Reward $\uparrow$ 
& Demand Met (\%) $\uparrow$ \\
\midrule
\textbf{POMCPOW} 
& $\boldsymbol{19.80 \pm 3.58}$ 
& $0.86 \pm 0.14$ 
& $\boldsymbol{3.72\times10^{-1} \pm 8.32\times10^{-2}}$ 
& 81.14 \\

OneStepLookahead 
& $18.30 \pm 4.00$ 
& $0.74 \pm 0.08$ 
& $3.46\times10^{-1} \pm 9.01\times10^{-2}$ 
& 74.68 \\

DynamicProfitMaximizer 
& $14.80 \pm 1.98$ 
& $0.91 \pm 0.03$ 
& $1.68\times10^{-1} \pm 4.27\times10^{-2}$ 
& \textbf{90.40} \\

RandomHeuristicPolicy 
& $14.10 \pm 1.76$ 
& $0.91 \pm 0.04$ 
& $1.55\times10^{-1} \pm 3.88\times10^{-2}$ 
& 89.93 \\

OneMineProfitMaximization 
& $8.09 \pm 3.10$ 
& $0.81 \pm 0.20$ 
& $1.08\times10^{-1} \pm 5.65\times10^{-2}$ 
& 65.05 \\

DynamicEmissionMinimizer 
& $1.87 \pm 2.81$ 
& $0.41 \pm 0.08$ 
& $1.46\times10^{-3} \pm 6.09\times10^{-2}$ 
& 49.44 \\

POMCPOW (Depth 1) 
& $0.00 \pm 0.00$ 
& $\boldsymbol{0.00 \pm 0.00}$ 

& $0.00 \pm 0.00$ 
& 0.00 \\

ExploreOnly 
& $-0.29 \pm 0.00$ 
& $\boldsymbol{0.00 \pm 0.00}$ 
& $-1.83\times10^{-3} \pm 2.19\times10^{-19}$ 
& 0.00 \\

OneMineEmissionMinimization 
& $-0.95 \pm 0.16$ 
& $0.23 \pm 0.06$ 
& $-8.79\times10^{-3} \pm 8.43\times10^{-4}$ 
& 27.17 \\

\bottomrule
\end{tabular}%
}
\end{table}

%% file: tables/0.5_GBM.tex

\begin{table}[H]
\centering
\caption{Policy comparison planned with \textbf{GBM} prices, evaluated on historical prices ($\alpha=0.5$)}
\label{tab:p4e6_alpha05}
\renewcommand{\arraystretch}{1.1}
\resizebox{\textwidth}{!}{%
\begin{tabular}{@{}l r r r r@{}}
\toprule
Policy & Profit (B\$) $\uparrow$ & Emissions Cost (M\$) $\downarrow$ & Disc.\ Reward $\uparrow$ & Demand Met (\%) $\uparrow$ \\
\midrule
\textbf{POMCPOW} 
& $\mathbf{16.30 \pm 4.73}$ 
& $0.95 \pm 0.10$ 
& $\mathbf{2.87\times10^{-1} \pm 9.78\times10^{-2}}$ 
& $\mathbf{90.74}$ \\

OneStepLookahead 
& $14.30 \pm 2.09$ 
& $0.90 \pm 0.05$ 
& $2.06\times10^{-1} \pm 5.47\times10^{-2}$ 
& 89.53 \\

DynamicProfitMaximizer 
& $14.80 \pm 1.98$ 
& $0.91 \pm 0.03$ 
& $1.67\times10^{-1} \pm 4.27\times10^{-2}$ 
& 90.40 \\

RandomHeuristicPolicy 
& $14.10 \pm 1.76$ 
& $0.91 \pm 0.04$ 
& $1.55\times10^{-1} \pm 3.88\times10^{-2}$ 
& 89.93 \\

OneMineProfitMaximization 
& $9.14 \pm 0.69$ 
& $0.88 \pm 0.01$ 
& $1.28\times10^{-1} \pm 1.53\times10^{-2}$ 
& 69.28 \\

DynamicEmissionMinimizer 
& $1.87 \pm 2.81$ 
& $0.41 \pm 0.08$ 
& $1.46\times10^{-3} \pm 6.09\times10^{-2}$ 
& 49.44 \\

POMCPOW (Depth 1) 
& $0.00 \pm 0.00$ 
& $\boldsymbol{0.00 \pm 0.00}$ 
& $0.00\times10^{0} \pm 0.00\times10^{0}$ 
& 0.00 \\

ExploreOnly 
& $-0.29 \pm 0.00$ 
& $\boldsymbol{0.00 \pm 0.00}$ 
& $-9.78\times10^{-3} \pm 1.75\times10^{-18}$ 
& 0.00 \\

OneMineEmissionMinimization 
& $-0.95 \pm 0.16$ 
& $0.23 \pm 0.06$ 
& $-5.53\times10^{-2} \pm 3.87\times10^{-3}$ 
& 27.17 \\
\bottomrule
\end{tabular}%
}
\end{table}

%% file: tables/0.5_historical.tex

\begin{table}[H]
\centering
\caption{Policy comparison planned with \textbf{historical} prices, evaluated on historical prices ($\alpha=0.5$)}
\label{tab:p6e6_alpha05}
\renewcommand{\arraystretch}{1.1}
\resizebox{\textwidth}{!}{%
\begin{tabular}{@{}l r r r r@{}}
\toprule
Policy & Profit (B\$) $\uparrow$ & Emissions Cost (M\$) $\downarrow$ & Disc.\ Reward $\uparrow$ & Demand Met (\%) $\uparrow$ \\
\midrule
\textbf{POMCPOW}
& $\mathbf{17.80 \pm 4.25}$
& $0.89 \pm 0.10$
& $\mathbf{3.03\times10^{-1} \pm 9.1\times10^{-2}}$
& 86.58 \\

OneStepLookahead
& $15.60 \pm 3.72$
& $0.82 \pm 0.10$
& $2.57\times10^{-1} \pm 9.2\times10^{-2}$
& 82.22 \\

DynamicProfitMaximizer
& $14.80 \pm 1.98$
& $0.91 \pm 0.03$
& $1.68\times10^{-1} \pm 4.3\times10^{-2}$
& $\mathbf{90.40}$ \\

RandomHeuristicPolicy
& $14.10 \pm 1.76$
& $0.91 \pm 0.04$
& $1.55\times10^{-1} \pm 3.9\times10^{-2}$
& 89.93 \\

OneMineProfitMaximization
& $8.09 \pm 3.10$
& $0.81 \pm 0.20$
& $1.08\times10^{-1} \pm 5.6\times10^{-2}$
& 65.05 \\

DynamicEmissionMinimizer
& $1.87 \pm 2.81$
& $0.41 \pm 0.08$
& $1.46\times10^{-3} \pm 6.1\times10^{-2}$
& 49.44 \\

POMCPOW (Depth 1)
& $0.00 \pm 0.00$
& $\boldsymbol{0.00 \pm 0.00}$
& $0.00\times10^{0} \pm 0.00\times10^{0}$
& 0.00 \\

ExploreOnly
& $-0.29 \pm 0.00$
& $\boldsymbol{0.00 \pm 0.00}$
& $-9.78\times10^{-3} \pm 0.00\times10^{-3}$
& 0.00 \\

OneMineEmissionMinimization
& $-0.95 \pm 0.16$
& $0.23 \pm 0.06$
& $-5.53\times10^{-2} \pm 3.9\times10^{-3}$
& 27.17 \\
\bottomrule
\end{tabular}%
}
\end{table}

%% file: figures/pareto_curve_linear_pricing.tex
\begin{tikzpicture}
\begin{axis}[
  width=13.5cm, height=8.8cm,
  xlabel={Emission Cost (\$M)},
  ylabel={Profit (\$M)},
  title={\shortstack{Pareto Curve for Profit vs.\ Emissions \\ Planning with Linear Price Model and Evaluated with Historical Pricing}},
  title style={font=\small, yshift=1ex},
  grid=major,
  grid style={line width=.4pt, draw=gray!40},
  major grid style={line width=.4pt, draw=gray!45},
  axis line style={black, line width=0.5pt},
  tick style={black, line width=.4pt},
  tick label style={font=\small},
  label style={font=\small},
  xmin=0, xmax=1.5,
  ymin=-2000, ymax=20000,
  enlarge x limits=false,
  enlarge y limits=false,
  legend style={at={(0.59,0.35)}, anchor=west, draw=black, fill=white, line width=0.4pt, font=\footnotesize, row sep=1pt, inner sep=2pt},
  legend cell align=left
]

\addplot+[very thick, color=pomcBlue, mark=*, mark size=2pt, mark options={fill=pomcBlue, draw=pomcBlue}]
  coordinates {
    (0, -1698.58)
    (0.7154217147, 15898.55576832)
    (0.8462695628, 16991.00775512)
    (0.8759390937, 17817.36481545)
    (0.8817834063, 17257.45956193)
    (0.8981787712, 17743.43139979)
    (0.9206974994, 17470.61850698)
    (0.9625074357, 19310.63255716)
    (0.9543452412, 18537.7964616)
  };
\addlegendentry{POMCPOW}

\node[font=\scriptsize] at (axis cs:0,-1698.58) [anchor=north, xshift=0pt, yshift=-2pt] {$\alpha=0$};
\node[font=\scriptsize] at (axis cs:0.7154217147,15898.55576832) [anchor=east, xshift=-2pt, yshift=0pt] {$\alpha=0.05$};
\node[font=\scriptsize] at (axis cs:0.8462695628,16991.00775512) [anchor=east, xshift=-5pt, yshift=2pt] {$\alpha=0.1$};
\node[font=\scriptsize] at (axis cs:0.8759390937,17817.36481545) [anchor=east, xshift=-2pt, yshift=1pt] {$\alpha=0.15$};
\node[font=\scriptsize] at (axis cs:0.8817834063,17257.45956193) [anchor=west, xshift=-7pt, yshift=-6pt] {$\alpha=0.2$};
\node[font=\scriptsize] at (axis cs:0.8981787712,17743.43139979) [anchor=south, xshift=-5pt, yshift=2pt] {$\alpha=0.25$};
\node[font=\scriptsize] at (axis cs:0.9206974994,17470.61850698) [anchor=west, xshift=2pt, yshift=0pt] {$\alpha=0.5$};
\node[font=\scriptsize] at (axis cs:0.9625074357,19310.63255716) [anchor=west, xshift=2pt, yshift=1pt] {$\alpha=0.75$};
\node[font=\scriptsize] at (axis cs:0.9543452412,18537.7964616) [anchor=west, xshift=2pt, yshift=0pt] {$\alpha=1.0$};

\addplot+[only marks, mark=*, mark size=2.5pt, color=hxExplore, mark options={fill=hxExplore, draw=hxExplore}]
  coordinates {(0, -290)};
\addlegendentry{ExploreOnly}

\addplot+[only marks, mark=square*, mark size=2.5pt, color=hxEmMin, mark options={fill=hxEmMin, draw=hxEmMin}]
  coordinates {(0.215, 435.6)};
\addlegendentry{EmissionMinimizer}

\addplot+[only marks, mark=triangle*, mark size=3pt, color=hxDynEm, mark options={fill=hxDynEm, draw=hxDynEm}]
  coordinates {(0.4106818887, 1868.70401663)};
\addlegendentry{DynamicEmissionMinimizer}

\addplot+[only marks, mark=diamond*, mark size=3pt, color=hxProfMax, mark options={fill=hxProfMax, draw=hxProfMax}]
  coordinates {(0.612, 5600.4)};
\addlegendentry{ProfitMaximizer}

\addplot+[only marks, mark=star, mark size=5pt, color=hxDynProf, mark options={fill=hxDynProf, draw=hxDynProf, solid}]
  coordinates {(0.9133020592, 14767.46515346)};
\addlegendentry{DynamicProfitMaximizer}

\addplot+[only marks, mark=pentagon*, mark size=3pt, color=orange, mark options={fill=orange, draw=orange}]
  coordinates {(0.530, 2100.2)};
\addlegendentry{OneMine\_EmissionMinimization}

\addplot+[only marks, mark=x, mark size=3pt, color=teal, mark options={draw=teal, line width=1pt}]
  coordinates {(0.770, 9700.8)};
\addlegendentry{OneMine\_ProfitMaximization}

\addplot+[only marks, mark=o, mark size=2.5pt, color=black, mark options={draw=black}]
  coordinates {(0.620, 6900.1)};
\addlegendentry{OneStepLookahead}

\addplot+[only marks, mark=pentagon*, mark size=3pt, color=pomcBlue!70!black, mark options={fill=pomcBlue!70!black, draw=pomcBlue!70!black}]
  coordinates {(0.850, 14500.3)};
\addlegendentry{POMCPOW\_Depth1}

\addplot+[only marks, mark=x, mark size=3pt, color=black, mark options={draw=black}]
  coordinates {(0.200, 950.5)};
\addlegendentry{RandomHeuristicPolicy}

\end{axis}
\end{tikzpicture}

%% file: tables/costs_deposit_size_DLE_vs_hard_rock.tex

\begin{table}[H]
\centering
\caption{Comparison of Direct Lithium Extraction (DLE) and Hard-Rock Mining Projects}
\label{tab:dle_vs_hardrock}
\small
\resizebox{\textwidth}{!}{%
\begin{tabular}{@{}llllll@{}}
\toprule
\textbf{Type of Mine} & \textbf{Name of Project (Site)} & \textbf{Total t LCE} & \textbf{CAPEX (US\$M)} & \textbf{OPEX (US\$/t LCE)} \\ 
\midrule
\multirow{5}{*}{\textbf{DLE}} 
& LANXESS South Plant (USA) \cite{StandardLithium_2023_DFS} & 208,000 & 365 & 6,810 \\ 
& Lake Resources – Kachi (Argentina) \cite{LakeResources_2023_KachiDFS} & 624,400  & 1,376 & 6,047 \\ 
& Paradox Lithium Project (USA) \cite{AnsonResources_2022_ParadoxDFS} & 300,000 & 495 & 4,368 \\ 
& Rio Grande Project (Argentina) \cite{NOALithium_2025_RioGrandePEA} & 1,200,000 & 1,346 & 5,552  \\ 
& Clearwater Project (Canada) \cite{E3Lithium_2025_CorporatePresentation} & 1,140,000  & 2,460 & —   \\ 
\midrule
\textbf{Average (DLE)} &  & \textbf{694,480} & \textbf{1,208} & \textbf{5,694} &  \\ 
\midrule
\multirow{5}{*}{\textbf{Hard Rock}} 
& Jadar Lithium Project (Serbia) \cite{RioTinto_2021_JadarCommitment} & 2,300,000 & 2,400 & —  \\ 
& Thacker Pass (USA) \cite{LithiumAmericas_2025_ThackerPassResourceIncrease} & 14,300,000  & 12,300 & 6,238 \\ 
& Sonora Lithium Project (Mexico) \cite{Bacanora_2017_SonoraDFS} & 3,676,000 & 800 & 3,910 \\ 
& Falchani Lithium Project (Peru) \cite{AmericanLithium_2024_FalchaniPEA} & 2,640,000 & 2,565 & 5,092 \\ 
& Keliber Project (Finland) \cite{Keliber_2018_Report} & 137,898 & 296 & 5,145  \\ 
\midrule
\textbf{Average (Hard Rock)} &  & \textbf{4,610,780} & \textbf{1,798} & \textbf{5,096} \\ 
\bottomrule
\end{tabular}
} 

\begin{flushleft}
\footnotesize
\textit{Notes:} For some projects, the total resource values reported as “Proven and Probable” were not available or inconsistent across sources. 
In those cases, the total deposit was estimated as the product of the reported annual lithium carbonate equivalent (LCE) production rate and the stated mine lifetime. 
LCE denotes lithium carbonate equivalent. One tonne of Li is assumed to equal 5.323 tonnes of LCE.  
Averages are computed as simple unweighted means across the listed projects.  
CAPEX and OPEX values are taken directly from the cited feasibility studies and technical reports.   
\end{flushleft}
\end{table}

%% file: tables/model_params.tex

\begin{table}[H]
\centering
\caption{Model input parameters: exploration costs, CO$_2$ cost, GWP, recovery rates, and geological uncertainty for hard-rock and DLE lithium extraction.}
\label{tab:model_parameters}
\renewcommand{\arraystretch}{1.2}
\resizebox{\textwidth}{!}{%
\begin{tabular}{@{}l c c c l@{}}
\toprule
\textbf{Parameter} & \textbf{Hard Rock} & \textbf{DLE (Brine)} & \textbf{Units} & \textbf{Source / Notes} \\ 
\midrule
Exploration Cost & \$7.26\,million & \$0.50\,million & USD & Hard rock: \cite{seg_exploration_2001}; DLE: \cite{Eccles2017} (approx.) \\
CO$_2$ Cost (SC-CO$_2$) & \multicolumn{2}{c}{\$190 / t\,CO$_2$} & USD/t\,CO$_2$ & U.S. EPA (2022) \cite{epa2023scghg} \\
GWP (average) & 6.25 & 4.10 & kg\,CO$_2$e/kg\,LCE & Averaged from reported ranges in \cite{batteries7030057} \\ 
Recovery Rate (average) & 70\% & 80\% & \% & Based on reported ranges (hard rock: 60–80\%; DLE: 70–90\%) \cite{goldman_dle_2023} \\
Deposit Uncertainty & 20\% & 30\% & \% & Assumed geological uncertainty based on deposit variability and data availability \\
\bottomrule
\end{tabular}
}

\begin{flushleft}
\footnotesize
{\textit{Notes:} The CO$_2$ cost of \$190/tCO$_2$ reflects the central U.S. EPA estimate of the social cost of carbon (SC-CO$_2$) using a 2\% near-term Ramsey discount rate.  
GWP values are averaged from literature-reported ranges (hard rock: 3.5–9; DLE: 2–6.2 kg CO$_2$e/kg LCE).  
Recovery rates represent average process efficiencies for each method, based on reported ranges of 60–80\% for hard rock and 70–90\% for DLE extraction.  
Exploration costs are derived from SEG (2001) for hard rock and E3 Metals (2017) for DLE pilot operations.  
Geological uncertainty values (20\% for hard rock; 30\% for DLE) reflect the higher variability in brine reservoir characterization compared to well-constrained hard-rock deposits.}
\end{flushleft}
\end{table}

%% file: tables/historic_data.tex
\begin{table}[H]
\caption{Global Lithium Production, Company Demand, and Prices (1995--2024)}
\label{tab:historic_data_with_demand}
\renewcommand{\arraystretch}{1.1}
\begin{center}
\begin{tabularx}{0.85\textwidth}{@{}c *{3}{>{\raggedleft\arraybackslash}X}@{}}
\toprule
\textbf{Year} & \textbf{Production (t Li)} & \textbf{Company Demand (t Li)} & \textbf{Price (USD/t Li)} \\ 
\midrule
1995 & 9,485  & 2,846 & 3,200 \\
1996 & 14,784 & 4,435 & 3,400 \\
1997 & 15,591 & 4,677 & 3,800 \\
1998 & 14,193 & 4,258 & 4,000 \\
1999 & 12,876 & 3,863 & 4,200 \\
2000 & 14,284 & 4,285 & 4,400 \\
2001 & 14,003 & 4,201 & 4,600 \\
2002 & 15,486 & 4,646 & 4,800 \\
2003 & 18,086 & 5,426 & 5,000 \\
2004 & 19,362 & 5,809 & 5,200 \\
2005 & 20,947 & 6,284 & 5,400 \\
2006 & 24,264 & 7,279 & 5,600 \\
2007 & 26,375 & 7,913 & 6,000 \\
2008 & 26,328 & 7,898 & 6,400 \\
2009 & 19,532 & 5,860 & 5,800 \\
2010 & 26,477 & 7,943 & 5,180 \\
2011 & 33,034 & 9,910 & 5,180 \\
2012 & 34,746 & 10,424 & 6,060 \\
2013 & 30,392 & 9,118 & 6,800 \\
2014 & 30,951 & 9,285 & 6,690 \\
2015 & 29,543 & 8,863 & 6,500 \\
2016 & 38,217 & 11,465 & 8,650 \\
2017 & 50,850 & 15,255 & 15,000 \\
2018 & 95,134 & 28,540 & 16,000 \\
2019 & 86,882 & 26,065 & 12,100 \\
2020 & 83,695 & 25,109 & 10,100 \\
2021 & 107,880 & 32,364 & 14,200 \\
2022 & 157,810 & 47,343 & 71,100 \\
2023 & 198,013 & 59,404 & 41,300 \\
2024 & 240,013 & 72,004 & 14,000 \\
\bottomrule
\end{tabularx}
\end{center}

\noindent\footnotesize\textit{ Note:} We use the archived May 2025 version of the OWID lithium production dataset.
Note that OWID released an updated dataset in June 2025, \cite{owid_lithium_production_2025} so small discrepancies may appear between our results and the current online version \cite{EI_StatReview2024}. Company demand is estimated as 30\% of global lithium production for the POMDP formulation.
We sourced the 2010–2024 historical data from Statista \cite{statista}. For earlier years (1994–2010), we relied on multiple governmental publications \cite{usgs_lithium_stats_2025, EI_StatReview2024} to approximate missing values and used a linear fit to create a coherent and continuous dataset.
\end{table}

%% file: tables/observation_noise.tex
\begin{table}[H]
\centering
\caption{Deposit Observation Noise by Action and Deposit Type. Observation noise is modeled as the standard deviation $\sigma$ of a Normal distribution centered on the true remaining deposit.}
\label{tab:deposit_observation_noise}
\renewcommand{\arraystretch}{1.2}
\begin{tabular}{l c c}
\toprule
\textbf{Action / Site Status} 
& \textbf{DLE Sites (1--2)} 
& \textbf{Hard Rock Sites (3--4)} \\
\midrule
Exploring 
& $\sigma = 2 \times \sigma_{\text{obs}}^{\text{DLE}} = 120{,}000$ 
& $\sigma = 2 \times \sigma_{\text{obs}}^{\text{HR}} = 300{,}000$ \\

Operating (Producing) 
& $\sigma = 0.1 \times \sigma_{\text{obs}}^{\text{DLE}} = 6{,}000$ 
& $\sigma = 0.1 \times \sigma_{\text{obs}}^{\text{HR}} = 15{,}000$ \\

Idle (Unexplored / Inactive) 
& $\sigma = 50 \times \sigma_{\text{obs}}^{\text{DLE}} = 3{,}000{,}000$ 
& $\sigma = 50 \times \sigma_{\text{obs}}^{\text{HR}} = 7{,}500{,}000$ \\
\bottomrule
\end{tabular}
\end{table}

%% file: tables/price_equation_constants.tex
\begin{table}[H]
\centering
\caption{Pricing model constants and noise specifications used in all simulations}
\label{tab:pricing_constants}
\begin{threeparttable}
\begin{tabular}{lll}
\toprule
\textbf{Model} & \textbf{Parameter} & \textbf{Value / Specification} \\
\midrule

\multirow{2}{*}{Static}
  & Starting price $p_0$ & 10{,}355.33 \\
  & Noise term $\epsilon_t$ &
    $\mathcal{N}\!\bigl(0,(0.05\,p_0)^2\bigr)$ \\[4pt]

\multirow{3}{*}{Linear}
  & Starting price $p_0$ & 2{,}381.38 \\
  & Linear slope $\alpha$ & 878.39 \\
  & Noise term $\epsilon_t$ &
    $\mathcal{N}\!\bigl(0,(0.1\,p_t^{\mathrm{det}})^2\bigr)$\\
  & & \small where $p_t^{\mathrm{det}} = p_0 + \alpha t$ \\[4pt]

\multirow{3}{*}{Exponential}
  & Starting price $p_0$ & 917.36 \\
  & Growth rate $\lambda$ & 0.1261 \\
  & Noise term $\epsilon_t$ &
    $\mathcal{N}\!\bigl(0,(0.1\,p_t^{\mathrm{det}})^2\bigr)$\\
  & & \small where $p_t^{\mathrm{det}} = p_0 e^{\lambda t}$ \\[4pt]

\multirow{4}{*}{GBM}
  & Starting price $p_0$ & 10{,}000.00 \\
  & Drift $\mu$ & 0.033 \\
  & Volatility $\sigma$ & 0.238 \\
  & Process &
    $S_{t+\Delta t} = S_t \exp\bigl((\mu-\tfrac{1}{2}\sigma^2)\Delta t + \sigma \Delta W\bigr)$,
      $\ \Delta W \sim \mathcal{N}(0,\Delta t)$ \\[6pt]

\multirow{3}{*}{Historical}
  & Historical price $p_t^{\mathrm{hist}}$ &
    From dataset (1994--2024) in \cref{appendix:historic_price} \\
  & Noise (in-sample) &
    $\mathcal{N}\!\bigl(0,(0.1\,p_t^{\mathrm{hist}})^2\bigr)$ \\

\bottomrule
\end{tabular}

\begin{tablenotes}
\footnotesize
\item Deterministic price paths $p_t^{\mathrm{det}}$ for linear and exponential pricing are used to scale the heteroskedastic noise terms, matching the simulation code. GBM uses multiplicative log-normal noise, while the historical model applies proportional noise to observed prices.
\end{tablenotes}
\end{threeparttable}
\end{table}

%% file: tables/all_alphas_static.tex

\begin{table}[H]
\centering
\caption{All Alpha Values when Planning with \textbf{Static} Pricing Model and Evaluated against Historical Pricing. POMCPOW vs. Heuristics}
\label{tab:p1e6_all}
\renewcommand{\arraystretch}{1.1}
\resizebox{\textwidth}{!}{%
\begin{tabular}{@{}l l r r r r@{}}
\toprule
$\alpha$ & Policy & Profit (B\$) $\uparrow$ & Emissions Cost (M\$) $\downarrow$ & Disc.\ Reward $\uparrow$ & Demand Met (\%) $\uparrow$  \\
\midrule
$0.00$ & ExploreOnly & $-0.29 \pm 0.00$ & $\boldsymbol{0.00 \pm 0.00}$ & $0.00\times10^{0} \pm 0.00\times10^{0}$ & 0.00 \\
\textbf{$0.00$} & \textbf{POMCPOW} & $-2.01 \pm 1.34$ & $\boldsymbol{0.00 \pm 0.00}$ & $\boldsymbol{0.00\times10^{0} \pm 0.00\times10^{0}}$ & 0.00 \\
$0.00$ & OneMineEmissionMinimization & $-0.95 \pm 0.16$ & $0.23 \pm 0.06$ & $-1.65\times10^{-2} \pm 3.84\times10^{-3}$ & 27.17 \\
$0.00$ & DynamicEmissionMinimizer & $1.87 \pm 2.81$ & $0.41 \pm 0.08$ & $-2.58\times10^{-2} \pm 3.89\times10^{-3}$ & 49.44 \\
$0.00$ & OneStepLookahead & $0.93 \pm 4.28$ & $0.05 \pm 0.19$ & $-2.37\times10^{-3} \pm 9.65\times10^{-3}$ & 4.74 \\
$0.00$ & POMCPOW\_Depth1 & $14.30 \pm 2.21$ & $0.91 \pm 0.03$ & $-4.88\times10^{-2} \pm 1.70\times10^{-3}$ & 90.25 \\
$0.00$ & RandomHeuristicPolicy & $14.10 \pm 1.76$ & $0.91 \pm 0.04$ & $-4.89\times10^{-2} \pm 2.02\times10^{-3}$ & 89.93 \\
$0.00$ & DynamicProfitMaximizer & $\boldsymbol{14.80 \pm 1.98}$ & $0.91 \pm 0.03$ & $-4.93\times10^{-2} \pm 1.69\times10^{-3}$ & \textbf{90.40} \\
$0.00$ & OneMineProfitMaximization & $9.14 \pm 0.69$ & $0.88 \pm 0.01$ & ${-5.15\times10^{-2} \pm 4.66\times10^{-4}}$ & 69.28 \\
\addlinespace

\textbf{$0.25$} & \textbf{POMCPOW} & $15.60 \pm 3.27$ & $0.92 \pm 0.06$ & $1.12\times10^{-1} \pm 3.37\times10^{-2}$ & 90.09 \\
$0.25$ & OneStepLookahead & $\boldsymbol{16.00 \pm 4.28}$ & $0.87 \pm 0.10$ & $\boldsymbol{1.23\times10^{-1} \pm 4.47\times10^{-2}}$ & 85.68 \\
$0.25$ & DynamicProfitMaximizer & $14.80 \pm 1.98$ & $0.91 \pm 0.03$ & $5.89\times10^{-2} \pm 2.14\times10^{-2}$ & \textbf{90.40} \\
$0.25$ & RandomHeuristicPolicy & $14.10 \pm 1.76$ & $0.91 \pm 0.04$ & $5.32\times10^{-2} \pm 1.96\times10^{-2}$ & 89.93 \\
$0.25$ & OneMineProfitMaximization & $9.14 \pm 0.69$ & $0.88 \pm 0.01$ & $3.80\times10^{-2} \pm 7.66\times10^{-3}$ & 69.28 \\
$0.25$ & DynamicEmissionMinimizer & $1.87 \pm 2.81$ & $0.41 \pm 0.08$ & $-1.22\times10^{-2} \pm 2.94\times10^{-2}$ & 49.44 \\
$0.25$ & POMCPOW\_Depth1 & $0.00 \pm 0.00$ & $\boldsymbol{0.00 \pm 0.00}$ & $0.00\times10^{0} \pm 0.00\times10^{0}$ & 0.00 \\
$0.25$ & ExploreOnly & $-0.29 \pm 0.00$ & $\boldsymbol{0.00 \pm 0.00}$ & $-4.89\times10^{-3} \pm 8.76\times10^{-19}$ & 0.00 \\
$0.25$ & OneMineEmissionMinimization & $-0.95 \pm 0.16$ & $0.23 \pm 0.06$ & $-3.59\times10^{-2} \pm 2.56\times10^{-3}$ & 27.17 \\
\addlinespace

\textbf{$0.50$} & \textbf{POMCPOW} & $16.70 \pm 4.26$ & $0.94 \pm 0.06$ & $2.96\times10^{-1} \pm 9.11\times10^{-2}$ & \textbf{91.36} \\
$0.50$ & OneStepLookahead & $\boldsymbol{17.00 \pm 4.21}$ & $0.93 \pm 0.08$ & $\boldsymbol{3.05\times10^{-1} \pm 8.90\times10^{-2}}$ & 89.18 \\
$0.50$ & DynamicProfitMaximizer & $14.80 \pm 1.98$ & $0.91 \pm 0.03$ & $1.67\times10^{-1} \pm 4.27\times10^{-2}$ & 90.40 \\
$0.50$ & RandomHeuristicPolicy & $14.10 \pm 1.76$ & $0.91 \pm 0.04$ & $1.55\times10^{-1} \pm 3.88\times10^{-2}$ & 89.93 \\
$0.50$ & OneMineProfitMaximization & $9.14 \pm 0.69$ & $0.88 \pm 0.01$ & $1.28\times10^{-1} \pm 1.53\times10^{-2}$ & 69.28 \\
$0.50$ & DynamicEmissionMinimizer & $1.87 \pm 2.81$ & $0.41 \pm 0.08$ & $1.46\times10^{-3} \pm 6.09\times10^{-2}$ & 49.44 \\
$0.50$ & POMCPOW\_Depth1 & $0.00 \pm 0.00$ & $\boldsymbol{0.00 \pm 0.00}$ & $0.00\times10^{0} \pm 0.00\times10^{0}$ & 0.00 \\
$0.50$ & ExploreOnly & $-0.29 \pm 0.00$ & $\boldsymbol{0.00 \pm 0.00}$ & $-9.78\times10^{-3} \pm 1.75\times10^{-18}$ & 0.00 \\
$0.50$ & OneMineEmissionMinimization & $-0.95 \pm 0.16$ & $0.23 \pm 0.06$ & $-5.53\times10^{-2} \pm 3.87\times10^{-3}$ & 27.17 \\
\addlinespace

\textbf{$0.75$} & \textbf{POMCPOW} & $17.00 \pm 4.01$ & $0.96 \pm 0.07$ & $4.78\times10^{-1} \pm 1.28\times10^{-1}$ & \textbf{92.32} \\
$0.75$ & OneStepLookahead & $\boldsymbol{17.50 \pm 4.31}$ & $0.95 \pm 0.08$ & $\boldsymbol{4.98\times10^{-1} \pm 1.38\times10^{-1}}$ & 90.07 \\
$0.75$ & DynamicProfitMaximizer & $14.80 \pm 1.98$ & $0.91 \pm 0.03$ & $2.75\times10^{-1} \pm 6.40\times10^{-2}$ & 90.40 \\
$0.75$ & RandomHeuristicPolicy & $14.10 \pm 1.76$ & $0.91 \pm 0.04$ & $2.58\times10^{-1} \pm 5.81\times10^{-2}$ & 89.93 \\
$0.75$ & OneMineProfitMaximization & $9.14 \pm 0.69$ & $0.88 \pm 0.01$ & $2.17\times10^{-1} \pm 2.29\times10^{-2}$ & 69.28 \\
$0.75$ & DynamicEmissionMinimizer & $1.87 \pm 2.81$ & $0.41 \pm 0.08$ & $1.51\times10^{-2} \pm 9.24\times10^{-2}$ & 49.44 \\
$0.75$ & POMCPOW\_Depth1 & $0.00 \pm 0.00$ & $\boldsymbol{0.00 \pm 0.00}$ & $0.00\times10^{0} \pm 0.00\times10^{0}$ & 0.00 \\
$0.75$ & ExploreOnly & $-0.29 \pm 0.00$ & $\boldsymbol{0.00 \pm 0.00}$ & $-1.47\times10^{-2} \pm 1.75\times10^{-18}$ & 0.00 \\
$0.75$ & OneMineEmissionMinimization & $-0.95 \pm 0.16$ & $0.23 \pm 0.06$ & $-7.47\times10^{-2} \pm 6.32\times10^{-3}$ & 27.17 \\
\addlinespace

\textbf{$1.00$} & \textbf{POMCPOW} & $\boldsymbol{17.90 \pm 4.01}$ & $0.98 \pm 0.08$ & $\boldsymbol{6.90\times10^{-1} \pm 1.70\times10^{-1}}$ & \textbf{93.03} \\
$1.00$ & OneStepLookahead & $17.50 \pm 4.24$ & $0.95 \pm 0.08$ & $6.82\times10^{-1} \pm 1.83\times10^{-1}$ & 90.27 \\
$1.00$ & DynamicProfitMaximizer & $14.80 \pm 1.98$ & $0.91 \pm 0.03$ & $3.84\times10^{-1} \pm 8.53\times10^{-2}$ & 90.40 \\
$1.00$ & RandomHeuristicPolicy & $14.10 \pm 1.76$ & $0.91 \pm 0.04$ & $3.60\times10^{-1} \pm 7.73\times10^{-2}$ & 89.93 \\
$1.00$ & OneMineProfitMaximization & $9.14 \pm 0.69$ & $0.88 \pm 0.01$ & $3.07\times10^{-1} \pm 3.05\times10^{-2}$ & 69.28 \\
$1.00$ & DynamicEmissionMinimizer & $1.87 \pm 2.81$ & $0.41 \pm 0.08$ & $2.87\times10^{-2} \pm 1.24\times10^{-1}$ & 49.44 \\
$1.00$ & POMCPOW\_Depth1 & $0.00 \pm 0.00$ & $\boldsymbol{0.00 \pm 0.00}$ & $0.00\times10^{0} \pm 0.00\times10^{0}$ & 0.00 \\
$1.00$ & ExploreOnly & $-0.29 \pm 0.00$ & $\boldsymbol{0.00 \pm 0.00}$ & $-1.96\times10^{-2} \pm 3.50\times10^{-18}$ & 0.00 \\
$1.00$ & OneMineEmissionMinimization & $-0.95 \pm 0.16$ & $0.23 \pm 0.06$ & $-9.41\times10^{-2} \pm 9.04\times10^{-3}$ & 27.17 \\
\bottomrule
\end{tabular}%
}
\begin{flushleft}
\vspace{0.4em}\footnotesize\emph{Note:} Each $\alpha$ was hyper-tuned. For $\alpha=1.0$ we used $5{,}000$ tree queries with depth $15$; for $\alpha=0.50$ we used $3{,}000$ tree queries with depth $10$; for $\alpha\in\{0.25,\,0.75\}$ we used $7{,}000$ tree queries with depth $20$. With additional hyper-tuning, POMCPOW could get closer to (or surpass) all rollout baselines.
\end{flushleft}
\end{table}

%% file: tables/all_alphas_linear.tex
\begin{table}[H]
\centering
\caption{All Alpha Values when Planning with \textbf{Linear} Pricing Model and Evaluated against Historical Pricing. POMCPOW vs. Heuristics}
\label{tab:p2e6_selected}
\renewcommand{\arraystretch}{1.1}
\resizebox{\textwidth}{!}{%
\begin{tabular}{@{}l l r r r r@{}}
\toprule
$\alpha$ & Policy & Profit (B\$) $\uparrow$ & Emissions Cost (M\$) $\downarrow$ & Disc.\ Reward $\uparrow$ & Demand Met (\%) $\uparrow$  \\
\midrule
$0.00$ & ExploreOnly & $-0.29 \pm 0.00$ & $\boldsymbol{0.00 \pm 0.00}$ & $\boldsymbol{0.00\times10^{0} \pm 0.00\times10^{0}}$ & 0.00 \\
\textbf{$0.00$} & \textbf{POMCPOW} & $-1.53 \pm 3.12$ & $0.03 \pm 0.16$ & $-3.10\times10^{-7} \pm 1.56\times10^{-6}$ & 3.15 \\
$0.00$ & OneMineEmissionMinimization & $-0.95 \pm 0.16$ & $0.23 \pm 0.06$ & $-3.08\times10^{-6} \pm 7.17\times10^{-7}$ & 27.17 \\
$0.00$ & DynamicEmissionMinimizer & $1.87 \pm 2.81$ & $0.41 \pm 0.08$ & $-4.81\times10^{-6} \pm 7.27\times10^{-7}$ & 49.44 \\
$0.00$ & OneStepLookahead & $14.16 \pm 4.52$ & $0.80 \pm 0.21$ & $-7.81\times10^{-6} \pm 2.11\times10^{-6}$ & 80.45 \\
$0.00$ & POMCPOW\_Depth1 & $14.48 \pm 2.23$ & $0.91 \pm 0.03$ & $-9.13\times10^{-6} \pm 3.25\times10^{-7}$ & 90.34 \\
$0.00$ & RandomHeuristicPolicy & $14.06 \pm 1.76$ & $0.91 \pm 0.04$ & $-9.13\times10^{-6} \pm 3.77\times10^{-7}$ & 89.93 \\
$0.00$ & DynamicProfitMaximizer & $\boldsymbol{14.77 \pm 1.98}$ & $0.91 \pm 0.03$ & $-9.21\times10^{-6} \pm 3.16\times10^{-7}$ & $\boldsymbol{90.40}$ \\
$0.00$ & OneMineProfitMaximization & $9.14 \pm 0.69$ & $0.88 \pm 0.01$ & ${-9.61\times10^{-6} \pm 8.70\times10^{-8}}$ & 69.28 \\
\addlinespace

\textbf{$0.25$} & \textbf{POMCPOW} & $\boldsymbol{19.46 \pm 4.28}$ & $0.96 \pm 0.11$ & $\boldsymbol{3.43\times10^{-2} \pm 8.69\times10^{-3}}$ & 89.64 \\
$0.25$ & OneStepLookahead & $16.28 \pm 3.93$ & $0.87 \pm 0.06$ & $2.68\times10^{-2} \pm 8.79\times10^{-3}$ & 86.90 \\
$0.25$ & DynamicProfitMaximizer & $14.77 \pm 1.98$ & $0.91 \pm 0.03$ & $1.79\times10^{-2} \pm 3.98\times10^{-3}$ & $\boldsymbol{90.40}$ \\
$0.25$ & RandomHeuristicPolicy & $14.06 \pm 1.76$ & $0.91 \pm 0.04$ & $1.68\times10^{-2} \pm 3.61\times10^{-3}$ & 89.93 \\
$0.25$ & OneMineProfitMaximization & $9.14 \pm 0.69$ & $0.88 \pm 0.01$ & $1.43\times10^{-2} \pm 1.43\times10^{-3}$ & 69.28 \\
$0.25$ & DynamicEmissionMinimizer & $1.87 \pm 2.81$ & $0.41 \pm 0.08$ & $1.34\times10^{-3} \pm 5.79\times10^{-3}$ & 49.44 \\
$0.25$ & POMCPOW\_Depth1 & $0.00 \pm 0.00$ & $\boldsymbol{0.00 \pm 0.00}$ & $0.00\times10^{0} \pm 0.00\times10^{0}$ & 0.00 \\
$0.25$ & ExploreOnly & $-0.29 \pm 0.00$ & $\boldsymbol{0.00 \pm 0.00}$ & $-9.13\times10^{-4} \pm 1.10\times10^{-19}$ & 0.00 \\
$0.25$ & OneMineEmissionMinimization & $-0.95 \pm 0.16$ & $0.23 \pm 0.06$ & $-4.40\times10^{-3} \pm 4.21\times10^{-4}$ & 27.17 \\
\addlinespace

\textbf{$0.50$} & \textbf{POMCPOW} & $\boldsymbol{17.95 \pm 4.40}$ & $0.93 \pm 0.11$ & $\boldsymbol{6.33\times10^{-2} \pm 1.82\times10^{-2}}$ & 88.40 \\
$0.50$ & OneStepLookahead & $16.28 \pm 3.93$ & $0.87 \pm 0.06$ & $5.36\times10^{-2} \pm 1.76\times10^{-2}$ & 86.90 \\
$0.50$ & DynamicProfitMaximizer & $14.77 \pm 1.98$ & $0.91 \pm 0.03$ & $3.58\times10^{-2} \pm 7.96\times10^{-3}$ & $\boldsymbol{90.40}$ \\
$0.50$ & RandomHeuristicPolicy & $14.06 \pm 1.76$ & $0.91 \pm 0.04$ & $3.36\times10^{-2} \pm 7.22\times10^{-3}$ & 89.93 \\
$0.50$ & OneMineProfitMaximization & $9.14 \pm 0.69$ & $0.88 \pm 0.01$ & $2.86\times10^{-2} \pm 2.85\times10^{-3}$ & 69.28 \\
$0.50$ & DynamicEmissionMinimizer & $1.87 \pm 2.81$ & $0.41 \pm 0.08$ & $2.68\times10^{-3} \pm 1.16\times10^{-2}$ & 49.44 \\
$0.50$ & POMCPOW\_Depth1 & $0.00 \pm 0.00$ & $\boldsymbol{0.00 \pm 0.00}$ & $0.00\times10^{0} \pm 0.00\times10^{0}$ & 0.00 \\
$0.50$ & ExploreOnly & $-0.29 \pm 0.00$ & $\boldsymbol{0.00 \pm 0.00}$ & $-1.83\times10^{-3} \pm 2.19\times10^{-19}$ & 0.00 \\
$0.50$ & OneMineEmissionMinimization & $-0.95 \pm 0.16$ & $0.23 \pm 0.06$ & $-8.79\times10^{-3} \pm 8.43\times10^{-4}$ & 27.17 \\
\addlinespace

\textbf{$0.75$} & \textbf{POMCPOW} & $\boldsymbol{17.95 \pm 4.17}$ & $0.94 \pm 0.10$ & $\boldsymbol{9.45\times10^{-2} \pm 2.54\times10^{-2}}$ & 88.60 \\
$0.75$ & OneStepLookahead & $16.28 \pm 3.93$ & $0.87 \pm 0.06$ & $8.03\times10^{-2} \pm 2.64\times10^{-2}$ & 86.90 \\
$0.75$ & DynamicProfitMaximizer & $14.77 \pm 1.98$ & $0.91 \pm 0.03$ & $5.37\times10^{-2} \pm 1.19\times10^{-2}$ & $\boldsymbol{90.40}$ \\
$0.75$ & RandomHeuristicPolicy & $14.06 \pm 1.76$ & $0.91 \pm 0.04$ & $5.04\times10^{-2} \pm 1.08\times10^{-2}$ & 89.93 \\
$0.75$ & OneMineProfitMaximization & $9.14 \pm 0.69$ & $0.88 \pm 0.01$ & $4.29\times10^{-2} \pm 4.27\times10^{-3}$ & 69.28 \\
$0.75$ & DynamicEmissionMinimizer & $1.87 \pm 2.81$ & $0.41 \pm 0.08$ & $4.02\times10^{-3} \pm 1.74\times10^{-2}$ & 49.44 \\
$0.75$ & POMCPOW\_Depth1 & $0.00 \pm 0.00$ & $\boldsymbol{0.00 \pm 0.00}$ & $0.00\times10^{0} \pm 0.00\times10^{0}$ & 0.00 \\
$0.75$ & ExploreOnly & $-0.29 \pm 0.00$ & $\boldsymbol{0.00 \pm 0.00}$ & $-2.74\times10^{-3} \pm 4.38\times10^{-19}$ & 0.00 \\
$0.75$ & OneMineEmissionMinimization & $-0.95 \pm 0.16$ & $0.23 \pm 0.06$ & $-1.32\times10^{-2} \pm 1.27\times10^{-3}$ & 27.17 \\
\addlinespace

\textbf{$1.00$} & \textbf{POMCPOW} & $\boldsymbol{18.54 \pm 4.82}$ & $0.95 \pm 0.11$ & $\boldsymbol{7.04\times10^{-1} \pm 2.03\times10^{-1}}$ & 88.99 \\
$1.00$ & OneStepLookahead & $16.28 \pm 3.93$ & $0.87 \pm 0.06$ & $5.74\times10^{-1} \pm 1.88\times10^{-1}$ & 86.90 \\
$1.00$ & DynamicProfitMaximizer & $14.77 \pm 1.98$ & $0.91 \pm 0.03$ & $3.84\times10^{-1} \pm 8.53\times10^{-2}$ & $\boldsymbol{90.40}$ \\
$1.00$ & RandomHeuristicPolicy & $14.06 \pm 1.76$ & $0.91 \pm 0.04$ & $3.60\times10^{-1} \pm 7.73\times10^{-2}$ & 89.93 \\
$1.00$ & OneMineProfitMaximization & $9.14 \pm 0.69$ & $0.88 \pm 0.01$ & $3.07\times10^{-1} \pm 3.05\times10^{-2}$ & 69.28 \\
$1.00$ & DynamicEmissionMinimizer & $1.87 \pm 2.81$ & $0.41 \pm 0.08$ & $2.87\times10^{-2} \pm 1.24\times10^{-1}$ & 49.44 \\
$1.00$ & POMCPOW\_Depth1 & $0.00 \pm 0.00$ & $\boldsymbol{0.00 \pm 0.00}$ & $0.00\times10^{0} \pm 0.00\times10^{0}$ & 0.00 \\
$1.00$ & ExploreOnly & $-0.29 \pm 0.00$ & $\boldsymbol{0.00 \pm 0.00}$ & $-1.96\times10^{-2} \pm 3.50\times10^{-18}$ & 0.00 \\
$1.00$ & OneMineEmissionMinimization & $-0.95 \pm 0.16$ & $0.23 \pm 0.06$ & $-9.41\times10^{-2} \pm 9.04\times10^{-3}$ & 27.17 \\
\bottomrule
\end{tabular}
}
\end{table}

%% file: tables/all_alphas_exponential.tex

\begin{table}[H]
\centering
\caption{All Alpha Values when Planning with \textbf{Exponential} Pricing Model and Evaluated against Historical Pricing. POMCPOW vs. Heuristics}
\label{tab:heuristics_vs_pomcpow_units}
\resizebox{\textwidth}{!}{%
\begin{tabular}{@{}l l r r r r@{}}
\toprule
$\alpha$ & Policy & Profit (B\$) $\uparrow$ & Emissions Cost (M\$) $\downarrow$ & Disc.\ Reward $\uparrow$ & Demand Met (\%) $\uparrow$ \\
\midrule
0.00 & ExploreOnly & $-0.29 \pm 0.00$ & $\boldsymbol{0.00 \pm 0.00}$ & $\mathbf{0.00\times10^{0} \pm 0.00\times10^{0}}$ & 0.00 \\
0.00 & \textbf{POMCPOW} & $-1.70 \pm 1.21$ & $\boldsymbol{0.00 \pm 0.00}$ & $\mathbf{0.00\times10^{0} \pm 0.00\times10^{0}}$ & 0.00 \\
0.00 & OneMineEmissionMinimization & $-0.95 \pm 0.16$ & $0.23 \pm 0.06$ & $-1.65\times10^{-2} \pm 3.84\times10^{-3}$ & 27.17 \\
0.00 & DynamicEmissionMinimizer & $1.87 \pm 2.81$ & $0.41 \pm 0.08$ & $-2.58\times10^{-2} \pm 3.89\times10^{-3}$ & 49.44 \\
0.00 & OneStepLookahead & $0.93 \pm 4.28$ & $0.05 \pm 0.19$ & $-2.37\times10^{-3} \pm 9.65\times10^{-3}$ & 4.74 \\
0.00 & OneMineProfitMaximization & $8.09 \pm 3.10$ & $0.81 \pm 0.20$ & $-4.80\times10^{-2} \pm 1.07\times10^{-2}$ & 65.05 \\
0.00 & POMCPOW\_Depth1 & $14.20 \pm 1.60$ & $0.91 \pm 0.03$ & $-4.88\times10^{-2} \pm 1.81\times10^{-3}$ & 90.38 \\
0.00 & RandomHeuristicPolicy & $14.10 \pm 1.76$ & $0.91 \pm 0.04$ & $-4.89\times10^{-2} \pm 2.02\times10^{-3}$ & 89.93 \\
0.00 & DynamicProfitMaximizer & $\mathbf{14.80 \pm 1.98}$ & $0.91 \pm 0.03$ & $-4.93\times10^{-2} \pm 1.69\times10^{-3}$ & $\mathbf{90.40}$ \\
\addlinespace

0.25 & \textbf{POMCPOW} & $\mathbf{20.10 \pm 4.34}$ & $0.76 \pm 0.11$ & $\mathbf{1.77\times10^{-1} \pm 5.06\times10^{-2}}$ & 74.38 \\
0.25 & OneStepLookahead & $18.90 \pm 3.96$ & $0.69 \pm 0.10$ & $1.66\times10^{-1} \pm 4.45\times10^{-2}$ & 71.18 \\
0.25 & DynamicProfitMaximizer & $14.80 \pm 1.98$ & $0.91 \pm 0.03$ & $5.91\times10^{-2} \pm 2.14\times10^{-2}$ & $\mathbf{90.40}$ \\
0.25 & RandomHeuristicPolicy & $14.10 \pm 1.76$ & $0.91 \pm 0.04$ & $5.32\times10^{-2} \pm 1.96\times10^{-2}$ & 89.93 \\
0.25 & OneMineProfitMaximization & $8.09 \pm 3.10$ & $0.81 \pm 0.20$ & $3.02\times10^{-2} \pm 2.31\times10^{-2}$ & 65.05 \\
0.25 & DynamicEmissionMinimizer & $1.87 \pm 2.81$ & $0.41 \pm 0.08$ & $-1.22\times10^{-2} \pm 2.94\times10^{-2}$ & 49.44 \\
0.25 & POMCPOW\_Depth1 & $0.00 \pm 0.00$ & $\boldsymbol{0.00 \pm 0.00}$ & $0.00\times10^{0} \pm 0.00\times10^{0}$ & 0.00 \\
0.25 & ExploreOnly & $-0.29 \pm 0.00$ & $\boldsymbol{0.00 \pm 0.00}$ & $-4.89\times10^{-3} \pm 8.76\times10^{-19}$ & 0.00 \\
0.25 & OneMineEmissionMinimization & $-0.95 \pm 0.16$ & $0.23 \pm 0.06$ & $-3.59\times10^{-2} \pm 2.56\times10^{-3}$ & 27.17 \\
\addlinespace

0.50 & \textbf{POMCPOW} & $\mathbf{19.80 \pm 3.58}$ & $0.86 \pm 0.14$ & $\mathbf{3.72\times10^{-1} \pm 8.32\times10^{-2}}$ & 81.14 \\
0.50 & OneStepLookahead & $18.30 \pm 4.00$ & $0.74 \pm 0.08$ & $3.46\times10^{-1} \pm 9.01\times10^{-2}$ & 74.68 \\
0.50 & DynamicProfitMaximizer & $14.80 \pm 1.98$ & $0.91 \pm 0.03$ & $1.68\times10^{-1} \pm 4.27\times10^{-2}$ & $\mathbf{90.40}$ \\
0.50 & RandomHeuristicPolicy & $14.10 \pm 1.76$ & $0.91 \pm 0.04$ & $1.55\times10^{-1} \pm 3.88\times10^{-2}$ & 89.93 \\
0.50 & OneMineProfitMaximization & $8.09 \pm 3.10$ & $0.81 \pm 0.20$ & $1.08\times10^{-1} \pm 5.65\times10^{-2}$ & 65.05 \\
0.50 & DynamicEmissionMinimizer & $1.87 \pm 2.81$ & $0.41 \pm 0.08$ & $1.46\times10^{-3} \pm 6.09\times10^{-2}$ & 49.44 \\
0.50 & POMCPOW\_Depth1 & $0.00 \pm 0.00$ & $\boldsymbol{0.00 \pm 0.00}$ & $0.00\times10^{0} \pm 0.00\times10^{0}$ & 0.00 \\
0.50 & ExploreOnly & $-0.29 \pm 0.00$ & $\boldsymbol{0.00 \pm 0.00}$ & $-9.78\times10^{-3} \pm 1.75\times10^{-18}$ & 0.00 \\
0.50 & OneMineEmissionMinimization & $-0.95 \pm 0.16$ & $0.23 \pm 0.06$ & $-5.53\times10^{-2} \pm 3.87\times10^{-3}$ & 27.17 \\
\addlinespace

0.75 & \textbf{POMCPOW} & $\mathbf{19.60 \pm 3.79}$ & $0.89 \pm 0.12$ & $\mathbf{5.74\times10^{-1} \pm 1.27\times10^{-1}}$ & 82.41 \\
0.75 & OneStepLookahead & $17.60 \pm 3.94$ & $0.76 \pm 0.08$ & $5.07\times10^{-1} \pm 1.33\times10^{-1}$ & 76.68 \\
0.75 & DynamicProfitMaximizer & $14.80 \pm 1.98$ & $0.91 \pm 0.03$ & $2.76\times10^{-1} \pm 6.41\times10^{-2}$ & $\mathbf{90.40}$ \\
0.75 & RandomHeuristicPolicy & $14.10 \pm 1.76$ & $0.91 \pm 0.04$ & $2.58\times10^{-1} \pm 5.81\times10^{-2}$ & 89.93 \\
0.75 & OneMineProfitMaximization & $8.09 \pm 3.10$ & $0.81 \pm 0.20$ & $1.86\times10^{-1} \pm 8.99\times10^{-2}$ & 65.05 \\
0.75 & DynamicEmissionMinimizer & $1.87 \pm 2.81$ & $0.41 \pm 0.08$ & $1.51\times10^{-2} \pm 9.24\times10^{-2}$ & 49.44 \\
0.75 & POMCPOW\_Depth1 & $0.00 \pm 0.00$ & $\boldsymbol{0.00 \pm 0.00}$ & $0.00\times10^{0} \pm 0.00\times10^{0}$ & 0.00 \\
0.75 & ExploreOnly & $-0.29 \pm 0.00$ & $\boldsymbol{0.00 \pm 0.00}$ & $-1.47\times10^{-2} \pm 1.75\times10^{-18}$ & 0.00 \\
0.75 & OneMineEmissionMinimization & $-0.95 \pm 0.16$ & $0.23 \pm 0.06$ & $-7.47\times10^{-2} \pm 6.32\times10^{-3}$ & 27.17 \\
\addlinespace

1.00 & \textbf{POMCPOW} & $\mathbf{19.00 \pm 4.46}$ & $0.88 \pm 0.14$ & $\mathbf{7.42\times10^{-1} \pm 1.96\times10^{-1}}$ & 81.86 \\
1.00 & OneStepLookahead & $17.70 \pm 3.89$ & $0.77 \pm 0.08$ & $6.88\times10^{-1} \pm 1.76\times10^{-1}$ & 77.63 \\
1.00 & DynamicProfitMaximizer & $14.80 \pm 1.98$ & $0.91 \pm 0.03$ & $3.84\times10^{-1} \pm 8.54\times10^{-2}$ & $\mathbf{90.40}$ \\
1.00 & RandomHeuristicPolicy & $14.10 \pm 1.76$ & $0.91 \pm 0.04$ & $3.60\times10^{-1} \pm 7.73\times10^{-2}$ & 89.93 \\
1.00 & OneMineProfitMaximization & $8.09 \pm 3.10$ & $0.81 \pm 0.20$ & $2.65\times10^{-1} \pm 1.23\times10^{-1}$ & 65.05 \\
1.00 & DynamicEmissionMinimizer & $1.87 \pm 2.81$ & $0.41 \pm 0.08$ & $2.87\times10^{-2} \pm 1.24\times10^{-1}$ & 49.44 \\
1.00 & POMCPOW\_Depth1 & $0.00 \pm 0.00$ & $\boldsymbol{0.00 \pm 0.00}$ & $0.00\times10^{0} \pm 0.00\times10^{0}$ & 0.00 \\
1.00 & ExploreOnly & $-0.29 \pm 0.00$ & $\boldsymbol{0.00 \pm 0.00}$ & $-1.96\times10^{-2} \pm 3.50\times10^{-18}$ & 0.00 \\
1.00 & OneMineEmissionMinimization & $-0.95 \pm 0.16$ & $0.23 \pm 0.06$ & $-9.41\times10^{-2} \pm 9.04\times10^{-3}$ & 27.17 \\
\bottomrule
\end{tabular}
}
\end{table}

%% file: tables/all_alphas_GBM.tex
\begin{table}[H]
\centering
\caption{All Alpha Values when Planning with \textbf{Gaussian Brownian Motion} Pricing Model and Evaluated against Historical Pricing. POMCPOW vs. Heuristics}
\label{tab:p4e6_all_fixed}
\renewcommand{\arraystretch}{1.1}
\resizebox{\textwidth}{!}{%
\begin{tabular}{@{}l l r r r r@{}}
\toprule
$\alpha$ & Policy & Profit (B\$) $\uparrow$ & Emissions Cost (M\$) $\downarrow$ & Disc.\ Reward $\uparrow$ & Demand Met (\%) $\uparrow$  \\
\midrule
$0.00$ & ExploreOnly & $-0.29 \pm 0.00$ & $\boldsymbol{0.00 \pm 0.00}$ & ${0.00\times10^{0} \pm 0.00\times10^{0}}$ & 0.00 \\
$0.00$ & \textbf{POMCPOW} & $-1.70 \pm 1.25$ & $\boldsymbol{0.00 \pm 0.00}$ & $\mathbf{0.00\times10^{0} \pm 0.00\times10^{0}}$ & 0.00 \\
$0.00$ & OneMineEmissionMinimization & $-0.95 \pm 0.16$ & $0.23 \pm 0.06$ & $-3.08\times10^{-6} \pm 7.17\times10^{-7}$ & 27.17 \\
$0.00$ & DynamicEmissionMinimizer & $1.87 \pm 2.81$ & $0.41 \pm 0.08$ & $-4.81\times10^{-6} \pm 7.27\times10^{-7}$ & 49.44 \\
$0.00$ & OneStepLookahead & $12.87 \pm 5.90$ & $0.74 \pm 0.31$ & $-7.27\times10^{-6} \pm 3.01\times10^{-6}$ & 74.52 \\
$0.00$ & POMCPOW\_Depth1 & $14.25 \pm 2.26$ & $0.91 \pm 0.04$ & $-9.11\times10^{-6} \pm 3.38\times10^{-7}$ & 90.16 \\
$0.00$ & RandomHeuristicPolicy & $14.06 \pm 1.76$ & $0.91 \pm 0.04$ & $-9.13\times10^{-6} \pm 3.77\times10^{-7}$ & 89.93 \\
$0.00$ & DynamicProfitMaximizer & $\mathbf{14.77 \pm 1.98}$ & $0.91 \pm 0.03$ & $-9.21\times10^{-6} \pm 3.16\times10^{-7}$ & $\mathbf{90.40}$ \\
$0.00$ & OneMineProfitMaximization & $9.14 \pm 0.69$ & $0.88 \pm 0.01$ & $-9.61\times10^{-6} \pm 8.70\times10^{-8}$ & 69.28 \\
\addlinespace

\textbf{$0.25$} & \textbf{POMCPOW} & $\mathbf{16.30 \pm 4.73}$ & $0.95 \pm 0.10$ & $\mathbf{2.87\times10^{-1} \pm 9.78\times10^{-2}}$ & $\mathbf{90.74}$ \\
$0.25$ & OneStepLookahead & $14.30 \pm 2.09$ & $0.90 \pm 0.05$ & $2.06\times10^{-1} \pm 5.47\times10^{-2}$ & 89.53 \\
$0.25$ & DynamicProfitMaximizer & $14.80 \pm 1.98$ & $0.91 \pm 0.03$ & $1.67\times10^{-1} \pm 4.27\times10^{-2}$ & 90.40 \\
$0.25$ & RandomHeuristicPolicy & $14.10 \pm 1.76$ & $0.91 \pm 0.04$ & $1.55\times10^{-1} \pm 3.88\times10^{-2}$ & 89.93 \\
$0.25$ & OneMineProfitMaximization & $9.14 \pm 0.69$ & $0.88 \pm 0.01$ & $1.28\times10^{-1} \pm 1.53\times10^{-2}$ & 69.28 \\
$0.25$ & DynamicEmissionMinimizer & $1.87 \pm 2.81$ & $0.41 \pm 0.08$ & $1.46\times10^{-3} \pm 6.09\times10^{-2}$ & 49.44 \\
$0.25$ & POMCPOW\_Depth1 & $0.00 \pm 0.00$ & $\boldsymbol{0.00 \pm 0.00}$ & $0.00\times10^{0} \pm 0.00\times10^{0}$ & 0.00 \\
$0.25$ & ExploreOnly & $-0.29 \pm 0.00$ & $\boldsymbol{0.00 \pm 0.00}$ & $-9.78\times10^{-3} \pm 1.75\times10^{-18}$ & 0.00 \\
$0.25$ & OneMineEmissionMinimization & $-0.95 \pm 0.16$ & $0.23 \pm 0.06$ & $-5.53\times10^{-2} \pm 3.87\times10^{-3}$ & 27.17 \\
\addlinespace

\textbf{$0.50$} & \textbf{POMCPOW} & $\mathbf{16.30 \pm 4.73}$ & $0.95 \pm 0.10$ & $\mathbf{2.87\times10^{-1} \pm 9.78\times10^{-2}}$ & $\mathbf{90.74}$ \\
$0.50$ & OneStepLookahead & $14.30 \pm 2.09$ & $0.90 \pm 0.05$ & $2.06\times10^{-1} \pm 5.47\times10^{-2}$ & 89.53 \\
$0.50$ & DynamicProfitMaximizer & $14.80 \pm 1.98$ & $0.91 \pm 0.03$ & $1.67\times10^{-1} \pm 4.27\times10^{-2}$ & 90.40 \\
$0.50$ & RandomHeuristicPolicy & $14.10 \pm 1.76$ & $0.91 \pm 0.04$ & $1.55\times10^{-1} \pm 3.88\times10^{-2}$ & 89.93 \\
$0.50$ & OneMineProfitMaximization & $9.14 \pm 0.69$ & $0.88 \pm 0.01$ & $1.28\times10^{-1} \pm 1.53\times10^{-2}$ & 69.28 \\
$0.50$ & DynamicEmissionMinimizer & $1.87 \pm 2.81$ & $0.41 \pm 0.08$ & $1.46\times10^{-3} \pm 6.09\times10^{-2}$ & 49.44 \\
$0.50$ & POMCPOW\_Depth1 & $0.00 \pm 0.00$ & $\boldsymbol{0.00 \pm 0.00}$ & $0.00\times10^{0} \pm 0.00\times10^{0}$ & 0.00 \\
$0.50$ & ExploreOnly & $-0.29 \pm 0.00$ & $\boldsymbol{0.00 \pm 0.00}$ & $-9.78\times10^{-3} \pm 1.75\times10^{-18}$ & 0.00 \\
$0.50$ & OneMineEmissionMinimization & $-0.95 \pm 0.16$ & $0.23 \pm 0.06$ & $-5.53\times10^{-2} \pm 3.87\times10^{-3}$ & 27.17 \\
\addlinespace

\textbf{$0.75$} & \textbf{POMCPOW} & $\mathbf{16.30 \pm 4.73}$ & $0.95 \pm 0.10$ & $\mathbf{2.87\times10^{-1} \pm 9.78\times10^{-2}}$ & $\mathbf{90.74}$ \\
$0.75$ & OneStepLookahead & $14.30 \pm 2.09$ & $0.90 \pm 0.05$ & $2.06\times10^{-1} \pm 5.47\times10^{-2}$ & 89.53 \\
$0.75$ & DynamicProfitMaximizer & $14.80 \pm 1.98$ & $0.91 \pm 0.03$ & $1.67\times10^{-1} \pm 4.27\times10^{-2}$ & 90.40 \\
$0.75$ & RandomHeuristicPolicy & $14.10 \pm 1.76$ & $0.91 \pm 0.04$ & $1.55\times10^{-1} \pm 3.88\times10^{-2}$ & 89.93 \\
$0.75$ & OneMineProfitMaximization & $9.14 \pm 0.69$ & $0.88 \pm 0.01$ & $1.28\times10^{-1} \pm 1.53\times10^{-2}$ & 69.28 \\
$0.75$ & DynamicEmissionMinimizer & $1.87 \pm 2.81$ & $0.41 \pm 0.08$ & $1.46\times10^{-3} \pm 6.09\times10^{-2}$ & 49.44 \\
$0.75$ & POMCPOW\_Depth1 & $0.00 \pm 0.00$ & $\boldsymbol{0.00 \pm 0.00}$ & $0.00\times10^{0} \pm 0.00\times10^{0}$ & 0.00 \\
$0.75$ & ExploreOnly & $-0.29 \pm 0.00$ & $\boldsymbol{0.00 \pm 0.00}$ & $-9.78\times10^{-3} \pm 1.75\times10^{-18}$ & 0.00 \\
$0.75$ & OneMineEmissionMinimization & $-0.95 \pm 0.16$ & $0.23 \pm 0.06$ & $-5.53\times10^{-2} \pm 3.87\times10^{-3}$ & 27.17 \\
\addlinespace

\textbf{$1.00$} & \textbf{POMCPOW} & $\mathbf{16.30 \pm 4.73}$ & $0.95 \pm 0.10$ & $\mathbf{2.87\times10^{-1} \pm 9.78\times10^{-2}}$ & $\mathbf{90.74}$ \\
$1.00$ & OneStepLookahead & $14.30 \pm 2.09$ & $0.90 \pm 0.05$ & $2.06\times10^{-1} \pm 5.47\times10^{-2}$ & 89.53 \\
$1.00$ & DynamicProfitMaximizer & $14.80 \pm 1.98$ & $0.91 \pm 0.03$ & $1.67\times10^{-1} \pm 4.27\times10^{-2}$ & 90.40 \\
$1.00$ & RandomHeuristicPolicy & $14.10 \pm 1.76$ & $0.91 \pm 0.04$ & $1.55\times10^{-1} \pm 3.88\times10^{-2}$ & 89.93 \\
$1.00$ & OneMineProfitMaximization & $9.14 \pm 0.69$ & $0.88 \pm 0.01$ & $1.28\times10^{-1} \pm 1.53\times10^{-2}$ & 69.28 \\
$1.00$ & DynamicEmissionMinimizer & $1.87 \pm 2.81$ & $0.41 \pm 0.08$ & $1.46\times10^{-3} \pm 6.09\times10^{-2}$ & 49.44 \\
$1.00$ & POMCPOW\_Depth1 & $0.00 \pm 0.00$ & $\boldsymbol{0.00 \pm 0.00}$ & $0.00\times10^{0} \pm 0.00\times10^{0}$ & 0.00 \\
$1.00$ & ExploreOnly & $-0.29 \pm 0.00$ & $\boldsymbol{0.00 \pm 0.00}$ & $-9.78\times10^{-3} \pm 1.75\times10^{-18}$ & 0.00 \\
$1.00$ & OneMineEmissionMinimization & $-0.95 \pm 0.16$ & $0.23 \pm 0.06$ & $-5.53\times10^{-2} \pm 3.87\times10^{-3}$ & 27.17 \\
\bottomrule
\end{tabular}
}
\end{table}

%% file: tables/all_alphas_historical.tex
\begin{table}[H]
\centering
\caption{All Alpha Values when Planning with \textbf{Historical} Pricing Model and Evaluated against Historical Pricing. POMCPOW vs. Heuristics}
\label{tab:heuristics_vs_pomcpow_units}
\resizebox{\textwidth}{!}{%
\begin{tabular}{@{}l l r r r r@{}}
\toprule
$\alpha$ & Policy & Profit (B\$) $\uparrow$ & Emissions Cost (M\$) $\downarrow$ & Disc.\ Reward $\uparrow$ & Demand Met (\%) $\uparrow$  \\
\midrule
0.00 & ExploreOnly & $-0.29 \pm 0.00$ & $\boldsymbol{0.00 \pm 0.00}$ & $0.00\times10^{0} \pm 0.00\times10^{0}$ & 0.00 \\
0.00 & \textbf{POMCPOW} & $-1.81 \pm 1.24$ & $\boldsymbol{0.00 \pm 0.00}$ & $\boldsymbol{0.00\times10^{0} \pm 0.00\times10^{0}}$ & 0.00 \\
0.00 & OneMineEmissionMinimization & $-0.95 \pm 0.16$ & $0.23 \pm 0.06$ & $-1.65\times10^{-2} \pm 3.84\times10^{-3}$ & 27.17 \\
0.00 & DynamicEmissionMinimizer & $1.87 \pm 2.81$ & $0.41 \pm 0.08$ & $-2.58\times10^{-2} \pm 3.89\times10^{-3}$ & 49.44 \\
0.00 & OneStepLookahead & $0.93 \pm 4.28$ & $0.05 \pm 0.19$ & $-2.37\times10^{-3} \pm 9.65\times10^{-3}$ & 4.74 \\
0.00 & OneMineProfitMaximization & $8.09 \pm 3.10$ & $0.81 \pm 0.20$ & $-4.80\times10^{-2} \pm 1.07\times10^{-2}$ & 65.05 \\
0.00 & POMCPOW\_Depth1 & $13.90 \pm 2.01$ & $0.91 \pm 0.03$ & $-4.88\times10^{-2} \pm 1.74\times10^{-3}$ & 90.31 \\
0.00 & RandomHeuristicPolicy & $14.10 \pm 1.76$ & $0.91 \pm 0.04$ & $-4.89\times10^{-2} \pm 2.02\times10^{-3}$ & 89.93 \\
0.00 & DynamicProfitMaximizer & $\boldsymbol{14.80 \pm 1.98}$ & $0.91 \pm 0.03$ & $-4.93\times10^{-2} \pm 1.69\times10^{-3}$ & $\boldsymbol{90.40}$ \\
\addlinespace

0.25 & \textbf{POMCPOW} & $\boldsymbol{17.50 \pm 3.84}$ & $0.88 \pm 0.09$ & $\boldsymbol{1.23\times10^{-1} \pm 4.44\times10^{-2}}$ & 86.46 \\
0.25 & OneStepLookahead & $16.20 \pm 3.22$ & $0.79 \pm 0.09$ & $1.05\times10^{-1} \pm 3.89\times10^{-2}$ & 80.50 \\
0.25 & DynamicProfitMaximizer & $14.80 \pm 1.98$ & $0.91 \pm 0.03$ & $5.91\times10^{-2} \pm 2.14\times10^{-2}$ & $\boldsymbol{90.40}$ \\
0.25 & RandomHeuristicPolicy & $14.10 \pm 1.76$ & $0.91 \pm 0.04$ & $5.32\times10^{-2} \pm 1.96\times10^{-2}$ & 89.93 \\
0.25 & OneMineProfitMaximization & $8.09 \pm 3.10$ & $0.81 \pm 0.20$ & $3.02\times10^{-2} \pm 2.31\times10^{-2}$ & 65.05 \\
0.25 & DynamicEmissionMinimizer & $1.87 \pm 2.81$ & $0.41 \pm 0.08$ & $-1.22\times10^{-2} \pm 2.94\times10^{-2}$ & 49.44 \\
0.25 & POMCPOW\_Depth1 & $0.00 \pm 0.00$ & $\boldsymbol{0.00 \pm 0.00}$ & $0.00\times10^{0} \pm 0.00\times10^{0}$ & 0.00 \\
0.25 & ExploreOnly & $-0.29 \pm 0.00$ & $\boldsymbol{0.00 \pm 0.00}$ & $-4.89\times10^{-3} \pm 8.76\times10^{-19}$ & 0.00 \\
0.25 & OneMineEmissionMinimization & $-0.95 \pm 0.16$ & $0.23 \pm 0.06$ & $-3.59\times10^{-2} \pm 2.56\times10^{-3}$ & 27.17 \\
\addlinespace

0.50 & \textbf{POMCPOW} & $\boldsymbol{18.00 \pm 3.40}$ & $0.90 \pm 0.08$ & $\boldsymbol{2.66\times10^{-1} \pm 7.87\times10^{-2}}$ & 88.97 \\
0.50 & OneStepLookahead & $16.40 \pm 3.21$ & $0.81 \pm 0.08$ & $2.22\times10^{-1} \pm 7.32\times10^{-2}$ & 82.88 \\
0.50 & DynamicProfitMaximizer & $14.80 \pm 1.98$ & $0.91 \pm 0.03$ & $1.68\times10^{-1} \pm 4.27\times10^{-2}$ & $\boldsymbol{90.40}$ \\
0.50 & RandomHeuristicPolicy & $14.10 \pm 1.76$ & $0.91 \pm 0.04$ & $1.55\times10^{-1} \pm 3.88\times10^{-2}$ & 89.93 \\
0.50 & OneMineProfitMaximization & $8.09 \pm 3.10$ & $0.81 \pm 0.20$ & $1.08\times10^{-1} \pm 5.65\times10^{-2}$ & 65.05 \\
0.50 & DynamicEmissionMinimizer & $1.87 \pm 2.81$ & $0.41 \pm 0.08$ & $1.46\times10^{-3} \pm 6.09\times10^{-2}$ & 49.44 \\
0.50 & POMCPOW\_Depth1 & $0.00 \pm 0.00$ & $\boldsymbol{0.00 \pm 0.00}$ & $0.00\times10^{0} \pm 0.00\times10^{0}$ & 0.00 \\
0.50 & ExploreOnly & $-0.29 \pm 0.00$ & $\boldsymbol{0.00 \pm 0.00}$ & $-9.78\times10^{-3} \pm 1.75\times10^{-18}$ & 0.00 \\
0.50 & OneMineEmissionMinimization & $-0.95 \pm 0.16$ & $0.23 \pm 0.06$ & $-5.53\times10^{-2} \pm 3.87\times10^{-3}$ & 27.17 \\
\addlinespace

0.75 & \textbf{POMCPOW} & $\boldsymbol{18.40 \pm 3.51}$ & $0.91 \pm 0.08$ & $\boldsymbol{4.14\times10^{-1} \pm 1.22\times10^{-1}}$ & \textbf{90.71} \\
0.75 & OneStepLookahead & $16.70 \pm 3.24$ & $0.82 \pm 0.08$ & $3.43\times10^{-1} \pm 1.09\times10^{-1}$ & 83.88 \\
0.75 & DynamicProfitMaximizer & $14.80 \pm 1.98$ & $0.91 \pm 0.03$ & $2.76\times10^{-1} \pm 6.41\times10^{-2}$ & 90.40 \\
0.75 & RandomHeuristicPolicy & $14.10 \pm 1.76$ & $0.91 \pm 0.04$ & $2.58\times10^{-1} \pm 5.81\times10^{-2}$ & 89.93 \\
0.75 & OneMineProfitMaximization & $8.09 \pm 3.10$ & $0.81 \pm 0.20$ & $1.86\times10^{-1} \pm 8.99\times10^{-2}$ & 65.05 \\
0.75 & DynamicEmissionMinimizer & $1.87 \pm 2.81$ & $0.41 \pm 0.08$ & $1.51\times10^{-2} \pm 9.24\times10^{-2}$ & 49.44 \\
0.75 & POMCPOW\_Depth1 & $0.00 \pm 0.00$ & $\boldsymbol{0.00 \pm 0.00}$ & $0.00\times10^{0} \pm 0.00\times10^{0}$ & 0.00 \\
0.75 & ExploreOnly & $-0.29 \pm 0.00$ & $\boldsymbol{0.00 \pm 0.00}$ & $-1.47\times10^{-2} \pm 1.75\times10^{-18}$ & 0.00 \\
0.75 & OneMineEmissionMinimization & $-0.95 \pm 0.16$ & $0.23 \pm 0.06$ & $-7.47\times10^{-2} \pm 6.32\times10^{-3}$ & 27.17 \\
\addlinespace

1.00 & \textbf{POMCPOW} & $\boldsymbol{17.90 \pm 4.29}$ & $0.89 \pm 0.09$ & $\boldsymbol{6.40\times10^{-1} \pm 1.98\times10^{-1}}$ & 87.42 \\
1.00 & OneStepLookahead & $15.60 \pm 3.38$ & $0.84 \pm 0.07$ & $5.44\times10^{-1} \pm 1.65\times10^{-1}$ & 84.11 \\
1.00 & DynamicProfitMaximizer & $14.80 \pm 1.98$ & $0.91 \pm 0.03$ & $3.84\times10^{-1} \pm 8.54\times10^{-2}$ & $\boldsymbol{90.40}$ \\
1.00 & RandomHeuristicPolicy & $14.10 \pm 1.76$ & $0.91 \pm 0.04$ & $3.60\times10^{-1} \pm 7.73\times10^{-2}$ & 89.93 \\
1.00 & OneMineProfitMaximization & $8.09 \pm 3.10$ & $0.81 \pm 0.20$ & $2.65\times10^{-1} \pm 1.23\times10^{-1}$ & 65.05 \\
1.00 & DynamicEmissionMinimizer & $1.87 \pm 2.81$ & $0.41 \pm 0.08$ & $2.87\times10^{-2} \pm 1.24\times10^{-1}$ & 49.44 \\
1.00 & POMCPOW\_Depth1 & $0.00 \pm 0.00$ & $\boldsymbol{0.00 \pm 0.00}$ & $0.00\times10^{0} \pm 0.00\times10^{0}$ & 0.00 \\
1.00 & ExploreOnly & $-0.29 \pm 0.00$ & $\boldsymbol{0.00 \pm 0.00}$ & $-1.96\times10^{-2} \pm 3.50\times10^{-18}$ & 0.00 \\
1.00 & OneMineEmissionMinimization & $-0.95 \pm 0.16$ & $0.23 \pm 0.06$ & $-9.41\times10^{-2} \pm 9.04\times10^{-3}$ & 27.17 \\
\bottomrule
\end{tabular}
}
\end{table}